\documentclass[10pt,twocolumn,letterpaper]{article}

\usepackage{cvpr}
\usepackage{times}
\usepackage{epsfig}
\usepackage{graphicx}
\usepackage{amsmath}
\usepackage{amssymb}
\usepackage{booktabs}
\usepackage{subcaption}
\usepackage{cite}
\usepackage{multirow}

\newcommand{\norm}[1]{\left\lVert #1 \right\rVert}

\usepackage[pagebackref=true,breaklinks=true,letterpaper=true,colorlinks,bookmarks=false]{hyperref}

\cvprfinalcopy 

\def\cvprPaperID{8601} 

\ifcvprfinal\fi

\begin{document}


\author{Bj{\"o}rn Browatzki and Christian Wallraven*\\
Dept. of Artificial Intelligence, Korea University, Seoul\\
{\tt\small browatbn@korea.ac.kr, wallraven@korea.ac.kr}
}

\title{3FabRec: Fast Few-shot Face alignment by Reconstruction}

\twocolumn[{%
\renewcommand\twocolumn[1][]{#1}%
\maketitle
\begin{center}
    \centering
    \includegraphics[width=1.0\linewidth]{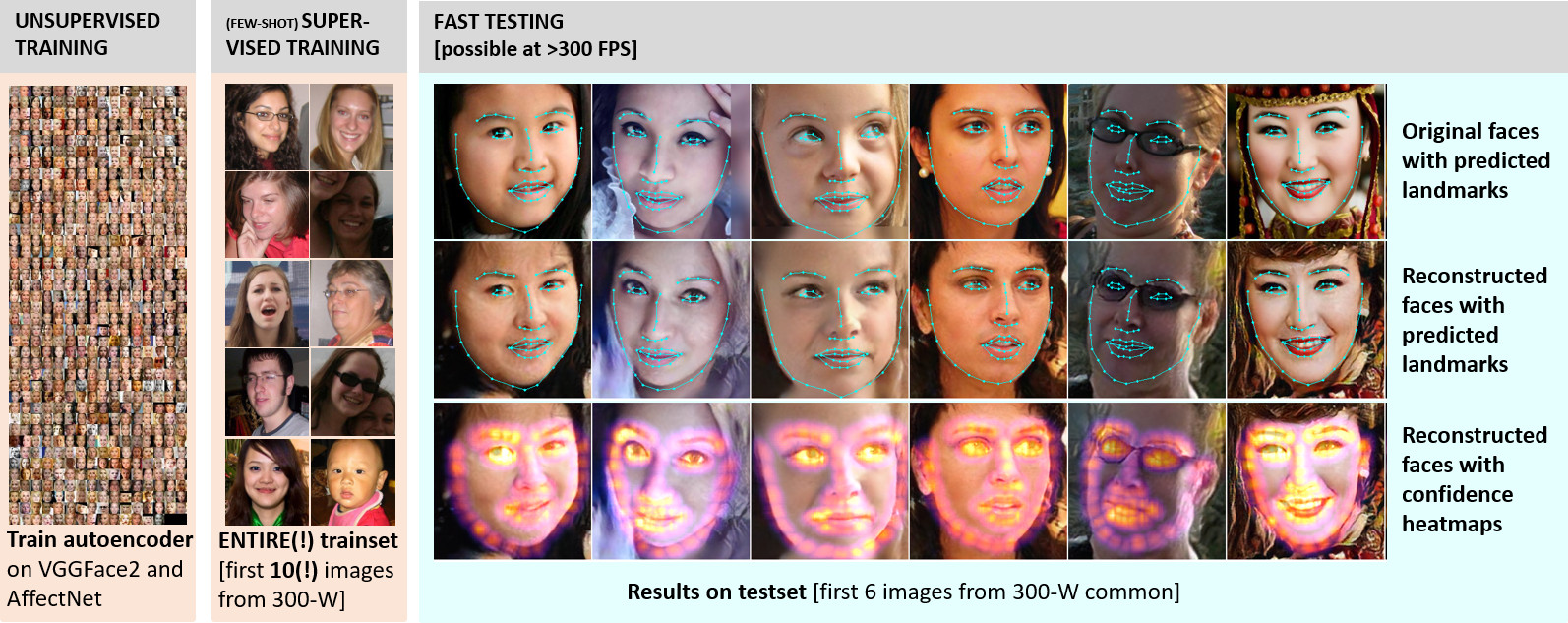}
    \vspace{-0.5cm}
    \captionof{figure}{\small The framework of 3FabRec consisting of: (leftmost box) a first, unsupervised training stage that trains a low-dimensional latent space via an adversarial (generative) autoencoder on a large dataset of unlabeled faces, (middle box) subsequent supervised training with few annotated faces. The rightmost box shows results from testing the framework trained on {\em only the 10 images of the middle box} with original faces (top row), reconstructed faces via the autoencoder (middle row), and confidence heatmaps (bottom row). 
}
    \label{fig:teaser}
\end{center}%
\vspace{0.3cm}
}]

\begin{abstract}\vspace{-0.35cm}

    Current supervised methods for facial landmark detection require a large amount of training data and may suffer from overfitting to specific datasets due to the massive number of parameters. We introduce a semi-supervised method in which the crucial idea is to first generate implicit face knowledge from the large amounts of unlabeled images of faces available today. In a first, completely unsupervised stage, we train an adversarial autoencoder to reconstruct faces via a low-dimensional face embedding. In a second, supervised stage, we interleave the decoder with transfer layers to retask the generation of color images to the prediction of landmark heatmaps. Our framework (3FabRec) achieves state-of-the-art performance on several common benchmarks and, most importantly, is able to maintain impressive accuracy on extremely small training sets down to as few as 10 images. As the interleaved layers only add a low amount of parameters to the decoder, inference runs at several hundred FPS on a GPU.

\end{abstract}

\vspace{-0.4cm}

\section{Introduction}\vspace{-0.2cm}

Accurate and robust localization of facial landmarks is a critical step in many existing face processing applications, including tracking, expression analysis, and face identification. Unique localization of such landmarks is severely affected by occlusions, partial face visibility, large pose variations, uneven illumination, or large, non-rigid deformations during more extreme facial expressions \cite{sagonas2013300,tresadern2011face}. These challenges have to be overcome in order to achieve a low landmark localization error, implying high robustness to appearance changes in faces while guaranteeing high localization accuracy for each landmark. 

The recent advances in deep learning techniques \cite{bodini2019review} coupled with the availability of large, annotated databases have allowed steady progress with localization accuracy on a typical benchmark increasing by 100\% (from \cite{xiong2013supervised} to \cite{Wu2018} - see below for more related work). Most approaches use a combination of highly-tuned, supervised learning schemes in order to achieve this performance and almost always are specifically optimized on the particular datasets that are tested, increasing the potential of overfitting to that dataset \cite{cawley2010over}. Similarly, it has been shown that annotations in datasets can be imprecise and inconsistent (e.g., \cite{Dong2018}).

Given that in addition to the existing annotated facial landmark datasets, there is an even larger number of datasets available for other tasks (face detection, face identification, facial expression analysis, etc.), it should be possible to leverage the implicit knowledge about face shape contained in this pool to both ensure better generalizability across datasets and easier and faster, few-shot training of landmark localization. Here, we present such a framework that is based on a two-stage architecture (3FabRec, see Figs.\ref{fig:teaser},\ref{fig:overview}): the key to the approach lies in the first, {\em unsupervised} stage, in which an (generative) adversarial autoencoder \cite{makhzani2015adversarial} is trained on a large dataset of faces that yields a low-dimensional embedding capturing "face knowledge" \cite{valentine2016face} from which it is able to reconstruct face images across a wide variety of appearances. With this embedding, the second, {\em supervised} stage then trains the landmark localization task on annotated datasets, in which the generator is retasked to predict the locations of a set of landmarks by generating probabilistic heatmaps \cite{Bulat2016}. This two-stage approach is a special case of semi-supervised learning \cite{kingma2014semi,zhu2017semi} and has been successful in other domains, including general network training \cite{hinton2006fast}, text classification \cite{howard2018universal} and translation \cite{devlin2018bert}, and visual image classification \cite{zhang2017split}.

In the current study, we show that our method is able to achieve state-of-the-art results running at $>$300 FPS on the standard benchmark datasets. Most importantly, it yields impressive localization performance already with {\em a few percent of the training data} - beating the leading scores in all cases and setting new standards for landmark localization from as few as {\em 10 images}. The latter result demonstrates that landmark knowledge has, indeed, been implicitly captured by the unsupervised pre-training. Additionally, the reconstructed autoencoder images are able to "explain away" extraneous factors (such as occlusions or make-up), yielding a best-fitting face shape for accurate localization and adding to the explainability of the framework. 

Source code is available at \url{https://github.com/browatbn2/3FabRec}.

\vspace{-0.0cm}

\section{Related Work}\vspace{0.1cm}
Before the advent of deep learning methods, explicitly-parametrized landmark models such as active shape \cite{cootes1992active}, active appearance \cite{cootes1998active} or cascade regression models \cite{xiong2013supervised,feng2015cascaded} provided the state-of-the-art in facial landmark detection. Current models using deep convolutional neural networks, however, quickly became the best-performing approaches, starting with deep alignment networks \cite{sun2013deep}, fully-convolutional networks \cite{liang2015unconstrained}, coordinate regression models \cite{Lv2017,trigeorgis2016mnemonic}, or multi-task learners \cite{ranjan2017hyperface}, with the deep networks being able to capture the pixel-to-landmark correlations across face appearance variations.

The recent related work in the context of our approach can be structured into supervised and semi-supervised approaches (for a recent, interesting unsupervised method - at lower performance levels - see \cite{thewlis2019unsupervised}).

\subsection{Supervised methods}
Several recent, well-performing supervised methods are based on heatmap regression, in which a deep network will infer a probabilistic heatmap for each of the facial landmarks with its corresponding maximum encoding the most likely location of that landmark \cite{liang2015unconstrained,Bulat2016,deng2019joint} - an approach we also follow here. In order to provide additional geometric constraints, extensions use an active-appearance-based model-fitting step based on PCA \cite{Merget2018},  explicit encoding of geometric information from the face boundary \cite{Wu2018}, or additional weighting from occlusion probabilities \cite{Meilu2019}. The currently best-performing method on many benchmarks uses a heatmap-based framework together with optimization of the loss function to foreground versus background pixels  \cite{Wang2019}. Such supervised methods will typically require large amounts of {\em labelled} training data in order to generalize across the variability in facial appearance (see \cite{Dapogny2019} for an architecture using high-resolution deep cascades that tries to address this issue).

\subsection{Semi-supervised methods}
In addition to changes to the network architecture, the issue of lack of training data and inconsistent labeling quality is addressed in semi-supervised models \cite{kingma2014semi,zhu2017semi} that augment the training process to make use of partially- or weakly-annotated data. Data augmentation based on landmark perturbation \cite{lv2016landmark} or from generating additional views from a 3D face model \cite{Zhu2016} can be applied to generate more robust pseudo landmark labels. \cite{Dong2018} uses constraints from temporal consistency of landmarks based on optic flow to enhance the training of the landmark detector - see also \cite{Zakharov2019}. In \cite{Zhang2014,ranjan2017hyperface}, multi-task frameworks are proposed in which attribute-networks tasked with predicting other facial attributes including pose and emotion are trained together with the landmark network, allowing for gradient transfer from one network to the other. Similar to this, \cite{Qian2019} show improvements using data augmentation with style-translated examples during training. In \cite{Dong2019}, a teacher-supervises-students (TS$^3$) framework is proposed in which a teacher is trained to filter student-generated landmark pseudolabels into "qualified" and "unqualified" samples, such that the student detectors can retrain themselves with better-quality data. Similarly, in \cite{Robinson2019}, a GAN framework produces "fake" heatmaps that the main branch of the network needs to discriminate, hence improving performance.

\vspace{-0.0cm}

\begin{figure*}[h!]
	\begin{center}
		\includegraphics[width=0.96\linewidth]{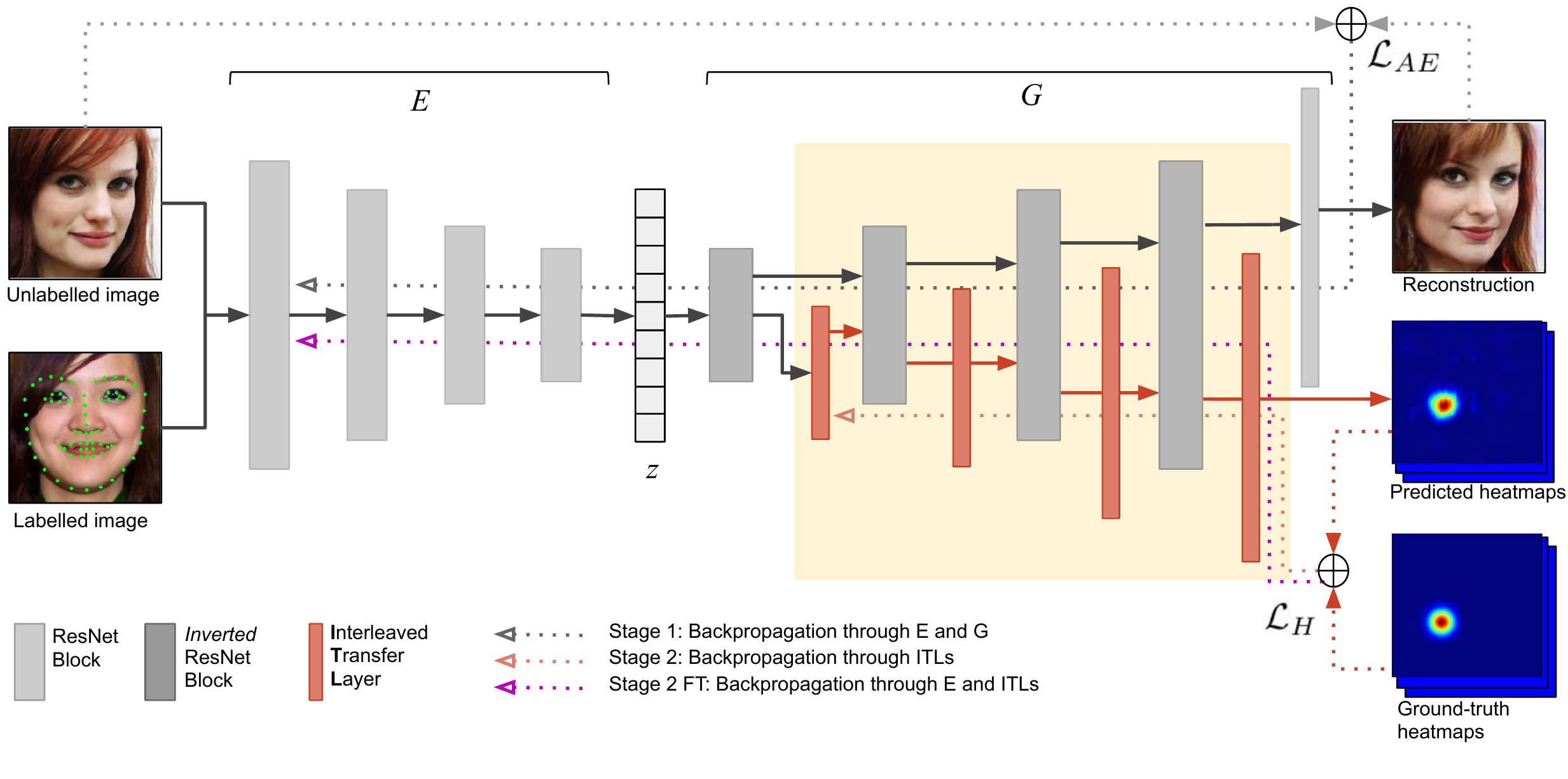}
	\end{center}
	\vspace{-0.8cm}
	\caption{\small Overview of the 3FabRec pipeline, including the architecture of the autoencoder, as well as training paths for unsupervised, supervised, and the fine-tuning stages (see text for more details).}
	\label{fig:overview}
	\vspace{-0.3cm}
\end{figure*}

\section{Methods}\label{sec:methods}\vspace{-0.0cm}
\subsection{Our approach}
Most of the semi-supervised approaches discussed above use data augmentation on the same dataset as done for testing. Our approach (see Figs. \ref{fig:teaser},\ref{fig:overview}) starts from an unsupervised method in which we leverage the {\em implicit} knowledge about face shape contained in large datasets of faces (such as used for face identification \cite{cao2018vggface2}). This knowledge is captured in a low-dimensional latent space of an autoencoder framework. Importantly, the autoencoder also has generative capabilities, i.e., it is tasked during training to reconstruct the face from the corresponding latent vector. This step is done because the following, supervised stage implements a hybrid reconstruction pipeline that uses the generator together with interleaved transfer layers to both reconstruct the face as well as probabilistic landmark heatmaps. Hence, the changes in the latent vector space will be mapped to the position of the landmarks trained on labeled datasets. Given that the first, unsupervised stage has already captured knowledge about facial appearance and face shape, this information will be quickly made explicit during the second, supervised stage allowing for generalization across multiple datasets and enabling low-shot and few-shot training.


\vspace{-0.0cm}\subsection{Unsupervised face representation}\vspace{-0.1cm} 
The unsupervised training step follows the framework of \cite{browatzki2019robust} in which an  adversarial autoencoder is trained through four loss functions  
balancing faithful image reconstruction with the generalizability and smoothness of the embedding space needed for the generation of novel faces. A reconstruction loss $\mathcal{L}_{rec}$ penalizes reconstruction errors through a pixel-based $L1$ error. An encoding feature loss $\mathcal{L}_{enc}$ \cite{goodfellow2014generative} ensures the creation of a smooth and continuous latent space. An adversarial feature loss  $\mathcal{L}_{adv}$ pushes the encoder $E$ and generator $G$ to produce reconstructions with high fidelity since training of generative models using only image reconstruction losses typically leads to blurred images. 

As the predicted landmark locations in our method follow directly from the locations of reconstructed facial elements, our main priority in training the autoencoder lies in the accurate reconstruction of such features. Thus, we trade some of the generative power against reconstruction accuracy by replacing the generative image loss, $\mathcal{L}_{gen}$, used in \cite{browatzki2019robust} with a new structural image loss  $\mathcal{L}_{cs}$. 

\vspace{-0.0cm}\paragraph{Structural image loss:} 
To penalize reconstructions that do not align facial structures well with input images, we add a structural image loss based on the SSIM\cite{wang2003multiscale} image similarity metric, which measures contrast $c(a,b)$ and correlation $s(a,b)$ between two image windows $a$ and $b$:

\begin{equation}
c(a,b) = \frac{2 \sigma_a \sigma_a + c}{\sigma_a^2 + \sigma_b^2 + c},~s(a,b) = \frac{\sigma_{ab}  + c/2}{\sigma_a \sigma_b + c/2}
\end{equation}

The values $\sigma_a$ and $\sigma_b$ denote intensity variances of windows $a$,$b$ and $\sigma_{ab}$ denotes their covariance.  The constant $c$ adds stability against small denominators. It is set to $c=255^{0.01}$ for images with 8-bit channels. The calculation is run for each $k\times k$ window across the images:

\begin{equation}
cs(x,y) = \frac{1}{|w|} \sum_w{c(x_w, y_w) s(x_w, y_w))}
\end{equation}

We obtain the structural image loss by evaluating $cs(x,y)$ with the original image and its reconstructions:

\vspace{-0.4cm}
\begin{equation}
\begin{split}
\mathcal{L}_{cs}(E,G) = \mathbb{E}_{x\sim p(x)}[ cs(x, G(E(x))]
\end{split}
\end{equation}

This loss improves the alignment of high-frequency image elements and imposes a penalty for high-frequency noise introduced by the adversarial image loss. Hence, $\mathcal{L}_{cs}$ also serves as a regularizer, stabilizing adversarial training.

\vspace{-0.2cm}\paragraph{Full autoencoder objective:} The final training objective is a weighted combination of all loss terms:
\begin{equation}
\begin{split}
\min_{E,G} \max_{D_{z}, D_{x}} &\mathcal{L}_{AE}(E,G,D_z,D_x)  = \\ 
& \lambda_{rec} \mathcal{L}_{rec}(E,G) + \lambda_{cs}\mathcal{L}_{cs}(E,G)  \\
&+ \lambda_{enc}\mathcal{L}_{enc}(E,D_z) + \lambda_{adv} \mathcal{L}_{adv}(E,G,D_x)
\end{split}
\end{equation}

We set $\lambda_{enc}$ and $\lambda_{adv}$ to $1.0$. $\lambda_{rec}$ and $\lambda_{cs}$ are selected so the corresponding loss terms yield similarly large values to each other, while at the same time ensuring a roughly 10 times higher weight in comparison to  $\lambda_{enc}$ and $\lambda_{adv}$ (given the range of loss terms, we set $\lambda_{rec}\approx 1.0$, $\lambda_{cs}\approx 60.0$).

\vspace{-0.0cm}\subsection{Supervised landmark discovery}\vspace{-0.0cm} 

For landmark detection, we are not primarily interested in producing a RGB image but rather an $L$-channel image containing landmark probability maps. This can be seen as a form of style transfer in which the appearance of the generated face is converted to a representation that allows us to read off landmark positions. Hence, information about face shape that was implicitly present in the generation of color images before is now made explicit.
Our goal is to create this transfer without losing the face knowledge distilled from the very large set of (unlabeled) images as the annotated datasets available for landmark prediction are only a fraction of that size and suffer from imprecise and inconsistent human annotations \cite{Dong2018}. For this, we introduce additional, interleaved transfer layers into the generator $G$.

\vspace{-0.2cm}\subsubsection{Interleaved transfer layers}\vspace{-0.1cm}
Training of landmark generation starts by freezing all parameters of the autoencoder. We then interleave the inverted ResNet layers of the generator with $3\times3$ convolutional layers.  Each of these Interleaved Transfer Layers (ITL) produces the same number of output channels as the original ResNet layer. Activations produced by a ResNet layer are transformed by these layers and fed into the next higher block. The last convolutional layer mapping to RGB images is replaced by a convolutional layer mapping to $L$-channel heatmap images ($L=$ number of landmarks to be predicted). This approach adds just enough flexibility to the generator to produce new heatmap outputs by re-using the pre-trained autoencoder weights.

Given an annotated face image $x$, the ground truth heatmap $H_i$ for each landmark $l_i \in \mathbb{R}^2$ consists of a 2D Normal distribution centered at $l_i$ and a standard deviation of $\sigma$. 
During landmark training and inference the activations $a_{1}$ produced by the first inverted ResNet layer for an encoded image $z=E(x)$ are passed to the first ITL layer. This will transfer the activations and feed it into the next, frozen inverted ResNet layer, such that the full cascade of ResNet and ITLs can reconstruct a landmark heatmap $\tilde{H}$.
The heatmap prediction loss $\mathcal{L}_{H}$ is defined as the $L2$ distance between predicted ($\tilde{H}$) and ground truth heatmap ($H$)

\vspace{-0.1cm}
\begin{equation}
\mathcal{L}_{H}(ITL)  = \mathbb{E}_{x\sim p(x)}[ \norm{H - ITL(a_1)}_2] \\ 
\end{equation}
The position of the landmark is $\tilde{l}_i=\underset{u,v} {\mathrm{argmax}} ~\tilde{H}_i(u,v)$.

\vspace{-0.5cm}\subsubsection{Encoder finetuning}\vspace{-0.1cm} 
Once training of the ITL layers reaches convergence we can perform an optional finetuning step. For this, the encoder $E$ is unfrozen so that ITL layers and encoder are optimized in tandem (see Fig.\ref{fig:overview}).
\vspace{-0.1cm}
\begin{equation}
\mathcal{L}_{H}(ITL)  \rightarrow \mathcal{L}_{H}(E, ITL) \\ 
\end{equation}
Since the updates are only based on landmark errors, this will push $E$ to encode input faces such that facial features are placed more precisely in reconstructed faces. At the same time, other attributes like gender, skin color, or illumination may be removed as these are not relevant for the landmark prediction task. Overfitting is avoided since the generator remains unchanged, which acts as a regularizer and limits the flexibility of the encoder.

\vspace{-0.1cm}

\section{Experiments\protect\footnote{For experiments on parameter tuning, cross-database results, and further ablation studies, see supplementary materials.}}

\vspace{-0.1cm}\subsection{Datasets}\vspace{-0.1cm}

\paragraph{VGGFace2 \& AffectNet} The dataset used for unsupervised training of the generative autoencoder combines two datasets: the VGGFace2 dataset \cite{cao2018vggface2}, which contains a total of 3.3 million faces collected with large variability in pose, age, illumination, and ethnicity in mind. From the full dataset, we removed faces with a height of less than 100 pixels resulting in 1.8 million faces (from 8631 unique identities). In addition, we add the AffectNet dataset \cite{Mollahosseini2017} that was designed for capturing a wide variety of facial expressions (thus providing additional variability in face shape), which contains 228k images, yielding a total of ~2.1M images for autoencoder training.

\vspace{-0.3cm}\paragraph{300-W} This dataset was assembled by \cite{Sagonas2016} from several sources, including LFPW \cite{belhumeur2013localizing}, AFW \cite{le2012interactive}, HELEN \cite{zhu2012face}, XM2VTS \cite{messer1999xm2vtsdb}, and own data and annotated semi-automatically with 68 facial landmarks. Using the established splits reported in \cite{ren2016face}, a total of 3,148 training images 
and 689 testing images were used in our experiments. The latter is further split into 554 images that constitute the {\em common} subset and a further 135 images that constitute the {\em challenging } subset. Additionally, 300-W contains 300 indoor and 300 outdoor images that define the {\em private testset} of the original 300-W challenge.

\vspace{-0.3cm}\paragraph{AFLW} This dataset \cite{koestinger2011annotated} contains 24,386 in-the-wild faces with an especially wide range of face poses (yaw angles from $-120^{\circ}$--$120^{\circ}$] and roll and pitch angles from $-90^{\circ}$--$90^{\circ}$). Following common convention, we used splits of 20,000 images for training and 4,386 for testing and trained with only 19 of the 21 annotated landmarks \cite{Lv2017}.

\vspace{-0.3cm}\paragraph{WFLW} The newest dataset in our evaluation protocol is from \cite{Wu2018} containing a total of 10,000 faces with a 7,500/2,500 train/test split. Images were sourced from the WIDER FACE dataset \cite{yang2017learning} and were manually annotated with a much larger number of 98 landmarks. The dataset contains different (partially overlapping) test subsets for evaluation where each subset varies in pose, expression, illumination, make-up, occlusion, or blur. 

\subsection{Experimental settings}

\subsubsection{Unsupervised autoencoder training}
\paragraph{Network architecture}
Our implementation is based on \cite{browatzki2019robust} which combines a standard ResNet-18 as encoder with an inverted ResNet-18 (first convolution layers in each block replaced by 4$\times$4 deconvolution layers) as decoder. Both encoder and decoder contain $\approx$10M parameters each. The encoded feature length is 99 dimensions. 

\vspace{-0.3cm}\paragraph{Training procedure}
We train the autoencoder for 50 epochs with an input/output size of $128\times128$ and a batchsize of 100 images. Upon convergence we add an additional ResNet layer to both the encoder and decoder and train for another 50 epochs with an image size of $256\times256$ to increase reconstruction fidelity with a batchsize of 50. We use the Adam optimizer \cite{kingma2014adam} ($\beta_1=0.0, \beta_2=0.999$) with a constant learning rate of $2\times10^{-5}$, which yielded robust settings for adversarial learning.
We apply data augmentations of random horizontal flipping ($p=0.5$), translation ($\pm4\%$) resizing ($94\%$ to $103\%$), rotation ($\pm 45^\circ$).

\vspace{-0.2cm}\subsubsection{Supervised landmark training}
Images are cropped using supplied bounding boxes and resized to $256\times256$. For creating ground truth heatmaps, we set $\sigma=7$.  In all experiments we train four ITL layers and generate landmark heatmaps of size $128x128$ by skipping the last generator layer (as detailed in \ref{sec:ablation}, higher generator layers contain mostly decorrelated local appearance information).
To train from the landmark dataset images, we apply data augmentations of random horizontal flipping ($p=0.5$), translation ($\pm4\%$) resizing ($\pm5\%$), rotation ($\pm 30^\circ$), and occlusion (at inference time no augmentation is performed).
The learning rate during ITL-only training is set to 0.001. During the optional finetuning stage we lower ITL learning rate to 0.0001 while keeping the encoder learning rate the same as during training (=$2\times10^{-5}$) and resetting Adam's $\beta_1$ to the default value of $0.9$.


\vspace{-0.2cm}\paragraph{Evaluation Metrics} Performance of facial landmark detection is reported here using normalized mean error (NME), failure rate (FR) at 10\% NME and area-under-the-curve (AUC) of the Cumulative Error Distribution (CED) curve. For 300-W and WFLW we use the distance between the outer eye-corners as the "inter-ocular" normalization. Due to the high number of profile faces in AFLW, errors are normalized using the width of the (square) bounding boxes following \cite{zhu2015face}.

\subsection{Qualitative results}
\begin{figure}
	\begin{center}
		\includegraphics[width=\linewidth]{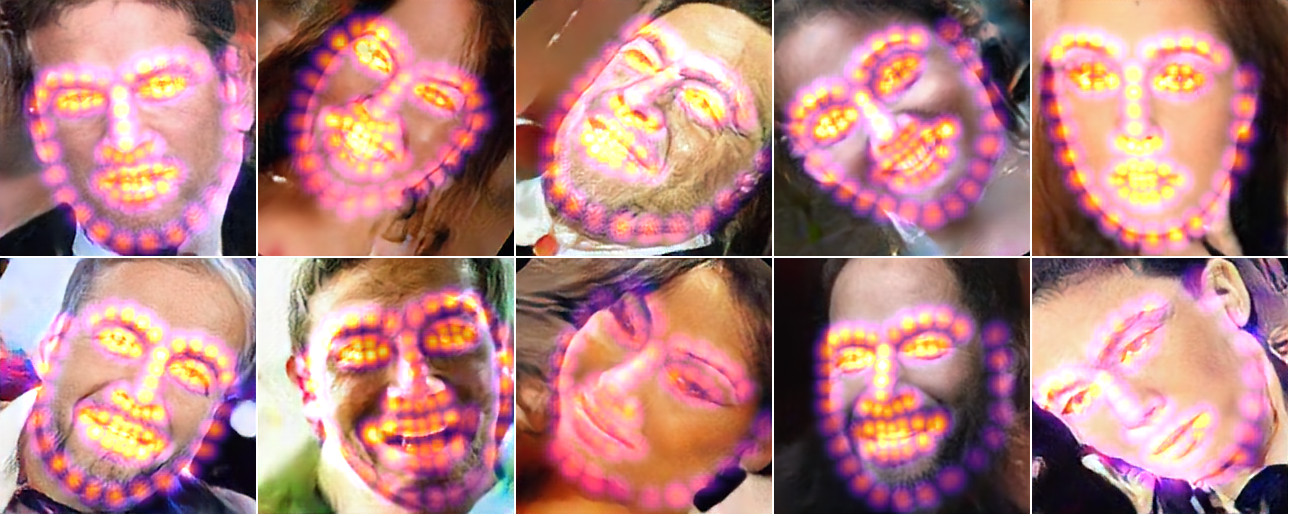}
	\end{center}
	\vspace{-0.5cm}
	\caption{\small Randomly-generated faces with overlaid generated landmark probability maps.}
	\label{fig:random}
	\vspace{-0.2cm}
\end{figure}

\begin{figure*}
	\begin{center}
		\includegraphics[width=\linewidth]{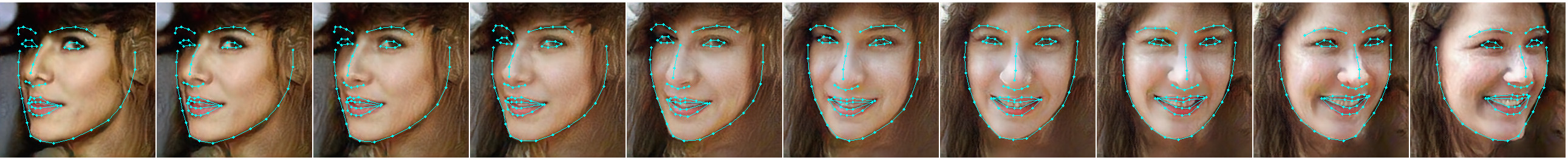}
	\end{center}
	\vspace{-0.5cm}
	\caption{\small Predicted landmarks of generated faces by interpolation between embedded feature vectors.}
	\label{fig:interp}
	\vspace{-0.2cm}
\end{figure*}

The trained generator is able to produce a wide range of realistic faces from a low-dimensional (99D) latent feature vector $z$ - this is shown in Fig.\ref{fig:random} with {\em randomly-generated} faces with overlaid, predicted landmark heatmaps. To achieve this, the model must have learned inherent information about the underlying structure of faces. We can further illustrate the implicit face shape knowledge by interpolating between face embeddings and observing that facial structures (such as mouth corners) in produced images are constructed in a highly consistent manner (see Fig. \ref{fig:interp} for a visualization). This leads to two insights: First, facial structures are actually encoded in the low-dimensional representation $z$. Second, this information can be transformed into 2D maps of pixel intensities (i.e., a color image) while maintaining high correlation with the originating encoding.

\begin{figure}
	\begin{center}
		\includegraphics[width=\linewidth]{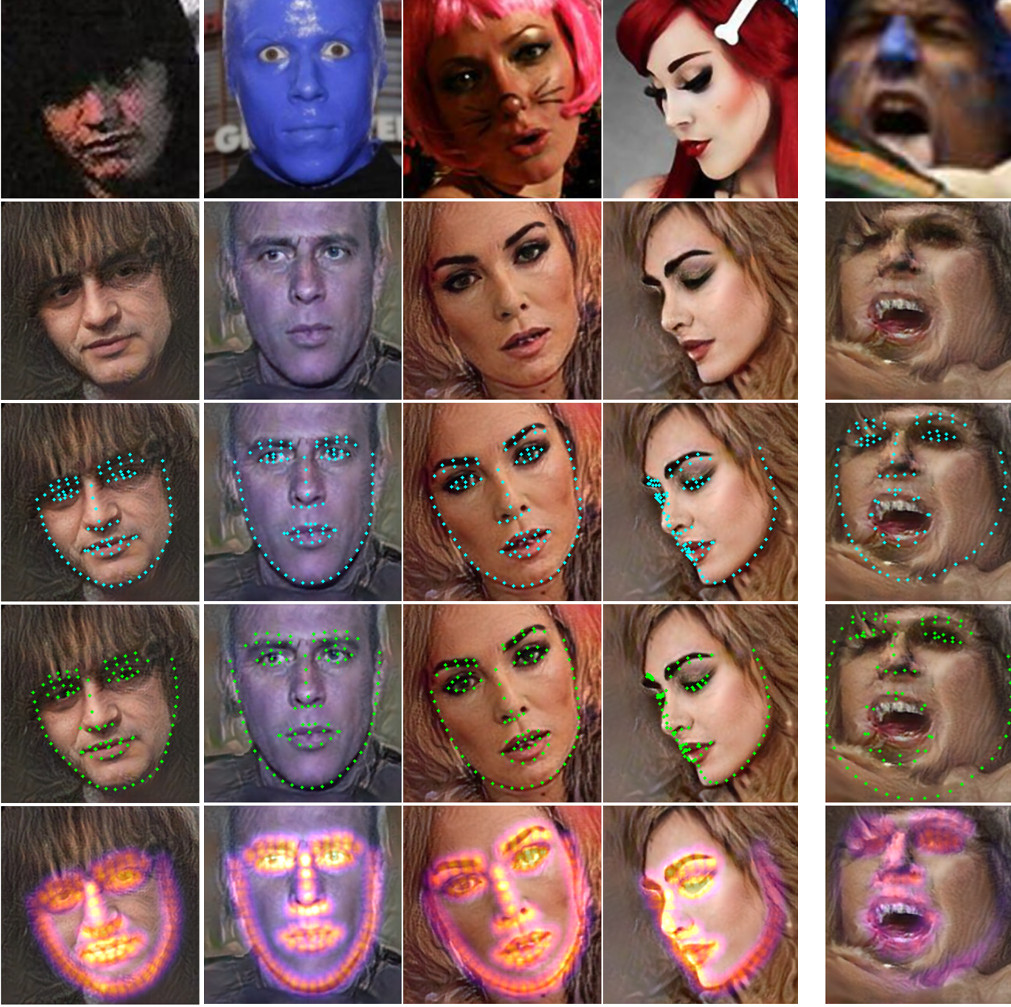}
	\end{center}
	\vspace{-0.5cm}
	\caption{\small 3FabRec results on challenging test examples from WFLW. Rows show the original, and the reconstruction itself, with predicted landmarks, with ground-truth landmarks, and with predicted landmark heatmaps, respectively. The fifth column illustrates a failure case. For more examples, see supplementary materials.}
	\label{fig:recocc}
	\vspace{-0.5cm}
\end{figure}

Further examples of the reconstruction quality on challenging images are shown in Fig. \ref{fig:recocc}. As can be seen, the pipeline will try to reconstruct a full face as much as possible given the input, removing occlusions and make-up and even "upsampling" the face (Fig. \ref{fig:recocc}, first column) in the process. This is because the databases for training the autoencoder contained mostly unoccluded and non-disguised faces at roughly similar resolutions. Additionally we note that the reconstructed faces will not necessarily preserve the identity as the goal of the fully-trained pipeline is to reconstruct the best-fitting face shape. Although our method is able to handle considerable variations in resolution  (Fig. \ref{fig:recocc}, first column), make-up  (Fig. \ref{fig:recocc}, second column), lighting  (Fig. \ref{fig:recocc}, third column), and pose  (Fig. \ref{fig:recocc}, fourth column), it does produce failed predictions in cases when these factors become too extreme, as shown in the fifth column of Fig. \ref{fig:recocc}. Landmark prediction, however, typically degrades gracefully in these cases as the confidence encoded in the heatmaps will also be low.

\vspace{-0.1cm}\subsection{Comparison with state-of-the-art}

Table \ref{tab:SOA} shows comparisons of our semi-supervised pipeline with state-of-the-art on the 300-W and the AFLW datasets using the full amount of training. We achieve top-2 accuracy on nearly all test sets with the exception of the common set from 300-W. This demonstrates that our framework is able to reach current levels of performance despite a much lighter, supervised training stage using only a few interleaved transfer layers on top of the generator pipeline.

The results in Table \ref{tab:AUCFR} for AUC and FR for the commonly-reported 300-W dataset demonstrate that our framework achieves the lowest failure rate of all methods (our FR=0.17 corresponds to only 1 image out of the full set that has large enough errors to count as a failure). At the same time, the AUC is in the upper range but not quite as good as that of \cite{Wu2018}, for example, which means that overall errors across landmarks are low, but more equally distributed compared to the top-performing methods.

\begin{table}
	\footnotesize
	\begin{center}
		\begin{tabular}{l| rr| rrr}
			\toprule
			
			& \multicolumn{2}{c}{\bfseries  AFLW} & \multicolumn{3}{c}{\bfseries  300-W}  \\
			\bfseries           Method &  \bf Full & \bf Frontal & \bfseries Com & \bfseries Chall. & \bfseries  Full \\
			
			\midrule
			
			SDM \cite{xiong2013supervised}                   &   4.05        &     2.94  &     5.57     &     15.40              &            7.52    \\
			LBF \cite{ren2014face} &  4.25        &     2.74  &        4.95     &    11.98               &            6.32    \\
			CFSS \cite{Zhu2015}     & 3.92   &  2.68  &    4.73     &    9.98        &            5.76    \\
			Two-Stage \cite{Lv2017} & 2.17 & - &     4.36 &     7.56    &  4.99  \\
			
			DSRN \cite{miao2018direct}   & 1.86 & - &        4.12     &     9.68               &            5.21    \\
			SBR \cite{dong2018supervision} &     2.14        &     2.07  &    3.28     &     7.58               &            4.10    \\
			SAN\cite{Dong2018} &     1.91    &   1.85 &     3.34     &     6.60               &            3.98    \\
			LAB \cite{Wu2018}    &   1.85   & 1.62 &   \bf{2.98}     &   \bf 5.19   & \bf 3.49    \\
			
			ODN \cite{Meilu2019}   &   \bf 1.63 & \bf 1.38 &      3.56     &     6.67               &            4.17    \\
			LaplaceKL (70K) \cite{Robinson2019} & 1.97 & - &  \underline{3.19}   &     6.87      &     3.91    \\
			\midrule
			3FabRec                 &   \underline{1.84}  & \underline{1.59} &     3.36     &     \underline{5.74}    &    \underline{3.82}    \\
			\midrule
			\bottomrule
		\end{tabular}	
	\end{center}
	\vspace{-0.5cm}
	\caption{\small Normalized mean error (\%) on 300-W dataset. Best results highlighted in bold, second best are underlined.}	
	\label{tab:SOA}
	\vspace{-0.0cm}\end{table}

\begin{table}
	\footnotesize
	\begin{center}
		\begin{tabular}{l r r}
			\toprule
			\bf Method & AUC & FR  \\
			\toprule
			M$^3$ CSR \cite{deng2016m3} & 47.52 &  5.5  \\
			CFSS \cite{zhu2015face}& 49.87 &  5.05  \\
			DenseReg+MDM \cite{alp2017densereg}& 52.19 &  3.67  \\
			JMFA \cite{deng2019joint}& 54.85 &  1.00  \\
			LAB \cite{Wu2018}& \bf 58.85 &  0.83  \\
			\midrule
			3FabRec & 54.61 &  \bf 0.17 \\
			
			\bottomrule
		\end{tabular}	
	\end{center}
	\vspace{-0.5cm}
	\caption{\small Area under the curve (AUC) and failure rate (FR in (\%) @0.1) on the 300-W testset.}	
	\label{tab:AUCFR}
	\vspace{-0.4cm}\end{table}

The NME results in Table \ref{tab:SOAWFLW} show that on the newest WFLW dataset, our approach performs at levels of the LAB method \cite{Wu2018} with most subsets, although we perform consistently below the current StyleAlign approach (SA, \cite{Qian2019} - note, however, that this approach could be easily implemented into our framework as well, which would allow us to disentangle the 99D-feature vector into style attributes \cite{browatzki2019robust} to generate augmented training data). The main reason for this is that WFLW contains much more heavy occlusions and extreme appearance changes compared to our training sets leading to more failure cases (see Fig.\ref{fig:recocc} fifth column).

\begin{table}
	\footnotesize
	\begin{center}
		\begin{tabular}{p{0.5cm}| p{1.55cm}| p{0.34cm} p{0.34cm}p{0.34cm}p{0.34cm}p{0.34cm}p{0.34cm}p{0.40cm}}
			\toprule
			\bf         & \bfseries  Method & \bf  Full  & \bf Pose & \bf Exp. & \bf Ill. & \bf Mk. Up & \bf Occ. & \bf Blur \\
			\toprule
			\multirow{4}{0.5cm}{NME (\%) }    
			& SDM \cite{xiong2013supervised} & 10.29 & 24.10 & 11.45 & 9.32 & 9.38 & 13.03 & 11.28 \\
			& CFSS \cite{Zhu2015} &  9.07 & 21.36 & 10.09 & 8.30 & 8.74 & 11.76 & 9.96 \\
			& DVLN \cite{Wu2015} &  6.08 & 11.54 & 6.78  & 5.73 & 5.98 & 7.33 & 6.88 \\
			& LAB \cite{Wu2018} & 5.27 & 10.24 & 5.51 & 5.23 & 5.15 & 6.79 & 6.32  \\
			& SAN \cite{Dong2018} & 5.22 & 10.39 & 5.71 & 5.19 & 5.49 & 6.83 & 5.80  \\
			& Wing \cite{Wu2018}  & 5.11 & 8.75 & 5.36 & 4.93 & 5.41 & 6.37 & 5.81  \\
			& SA \cite{Qian2019}  & \bf 4.39 & \bf 8.24 & \bf 4.68 & \bf 4.24 & \bf 4.37 & \bf 5.60 & \bf 4.86  \\
			\cmidrule{2-9}
			& 3FabRec             & 5.62 & 10.23 &6.09  &5.55 & 5.68 & 6.92 & 6.38  \\
			\cmidrule{2-9}
			\toprule
			\multirow{4}{0.5cm}{FR @0.1 (\%)}    
			& SDM \cite{xiong2013supervised} & 29.40 & 84.36 & 33.44 & 26.22 & 27.67&  41.85 & 35.32\\
			& CFSS \cite{Zhu2015} & 20.56 & 66.26 & 23.25 & 17.34 & 21.84 & 32.88 & 23.67 \\  
			& DVLN \cite{Wu2015}  & 10.84 & 46.93 & 11.15 & 7.31 & 11.65 & 16.30 & 13.71  \\
			& LAB \cite{Wu2018}   & 7.56 & 28.83 & 6.37 & 6.73 & 7.77 & 13.72 & 10.74  \\
			& SAN \cite{Dong2018} & 6.32 & 27.91 & 7.01 & 4.87 & 6.31 & 11.28 & 6.60  \\
			& Wing \cite{Wu2018}  & 6.00 & 22.70 & 4.78 & 4.30 & 7.77 & 12.50 & 7.76  \\
			& SA \cite{Qian2019}  & \bf 4.08 & \bf 18.10 & \bf 4.46 & \bf 2.72 & \bf 4.37 & \bf 7.74 & \bf 4.40  \\
			\cmidrule{2-9}
			& 3FabRec             & 8.28 & 34.35 & 8.28 &6.73 &10.19 &15.08 & 9.44  \\
			\toprule
			\multirow{4}{0.5cm}{AUC @0.1}        
			& SDM \cite{xiong2013supervised} & 0.300 & 0.023 & 0.229 & 0.324 & 0.312 & 0.206 & 0.239\\
			& CFSS \cite{Zhu2015} & 0.366 & 0.063 & 0.316 & 0.385 & 0.369 & 0.269 & 0.304 \\
			& DVLN \cite{Wu2015}  & 0.455 & 0.147 & 0.389 & 0.474 & 0.449 & 0.379 & 0.397 \\
			& LAB \cite{Wu2018}   & 0.532 & 0.235 & 0.495 & 0.543 & 0.539 & 0.449 & 0.463  \\
			& SAN \cite{Dong2019} & 0.536 & 0.236 & 0.462 & 0.555 & 0.522 & 0.456 & 0.493  \\
			& Wing \cite{Wu2018}  & 0.534 & \bf 0.310 & 0.496 & 0.541 & 0.558 & 0.489 & 0.492  \\
			& SA \cite{Qian2019}  & \bf 0.591 & 0.311 & \bf 0.549 & \bf 0.609 & \bf 0.581 & \bf 0.516 & \bf 0.551 \\
			\cmidrule{2-9}
			& 3FabRec             & 0.484 & 0.192 &0.448 &0.496 &0.473 & 0.398 & 0.434 \\
			
			\bottomrule
		\end{tabular}	
	\end{center}
	\vspace{-0.5cm}
	\caption{\small Evaluation results on WFLW dataset.}	
	\label{tab:SOAWFLW}
	\vspace{-0.3cm}\end{table}

\begin{table*}[ht]
	\footnotesize
	\begin{center}
		\begin{tabular}{p{1.47cm}| 
				p{0.31cm}p{0.31cm}p{0.31cm}| p{0.31cm}p{0.31cm}p{0.31cm} | p{0.31cm}p{0.31cm}p{0.31cm} | p{0.31cm}p{0.31cm}p{0.31cm} | p{0.31cm}p{0.31cm}p{0.31cm} | p{0.31cm}p{0.31cm}p{0.31cm} | p{0.31cm}p{0.31cm}p{0.31cm}}
			\toprule
			\multicolumn{22}{c}{\bfseries  300-W dataset}  \\
			\toprule
			\bfseries Method     &  \multicolumn{21}{c}{\bfseries  Training set size}  \\
			&    \multicolumn{3}{c}{100\%} &   \multicolumn{3}{c}{20\%}          & \multicolumn{3}{c}{10\%}             &   \multicolumn{3}{c}{5\%}        & \multicolumn{3}{c}{50 (1.5\%)} &  \multicolumn{3}{c}{10 (0.3\%)} & \multicolumn{3}{c}{1 (0.003\%)} \\
			\toprule
			RCN$^+$ \cite{Honari2018}   &  4.20&7.78&4.90             &  -   &9.56 & 5.88   &     -&10.35&6.32       &   -&15.54&7.22    &  -&-&-      &  -&-&-         &   -&-&-            \\
			RCN$^+$ \cite{Honari2018}$^\dag$ &  3.00&\bf 4.98&\bf 3.46        &  -   &\bf 6.12 &4.15   &     -&\bf 6.63& \bf 4.47       &   -&9.95&5.11    &  -&-&-      &  -&-&-         &   -&-&-            \\
			SA \cite{Qian2019}          &  3.21&6.49&3.86             &  3.85&-&-        &  4.27&-&-              &  6.32&-&-      &  &&         &     && &  &&           \\
			TS$^3$ \cite{Dong2019}      &  \bf 2.91& 5.9 & 3.49    &  4.31&7.97&5.03   &  4.67& 9.26&5.64        &  -&-&-         & -&-&-      &  -&-&-         &   -&-&-            \\
			\midrule
			3FabRec                     &  3.36& 5.74&3.82        & \bf 3.76& 6.53 & 4.31   & \bf 3.88& 6.88 & \bf 4.47    &  \bf 4.22& \bf 6.95& \bf 4.75    & \bf 4.55& \bf 7.39 & \bf 5.10    &  \bf 4.96& \bf 8.29& \bf 5.61      &    \bf 8.45&\bf 15.84&\bf 9.92             \\
			\midrule
			\bottomrule
		\end{tabular}	
	\end{center}
	\vspace{-0.5cm}
	\caption{\small NME (\%) with reduced training sets on 300-W. $^\dag$ RCN$^+$ reports errors normalized by eye-center distance - for better comparison values were rescaled by the known ratios of inter-ocular to inter-pupil distances, "-" denotes values not reported.}	
	\label{tab:fewshot300}
	\vspace{0.1cm}
\end{table*}

\subsection{Limited training data and few-shot learning }
Tables \ref{tab:fewshot300}, \ref{tab:fewshotAFLW}, \ref{tab:fewshotWFLW} showcase the central result of our framework: when training on only parts of the training set, 3FabRec can beat the published benchmark performance values. 

\vspace{-0.3cm}\paragraph{300-W} Table \ref{tab:fewshot300} shows that performance is comparable to that of 2-year-old approaches trained on the full dataset (cf. Table \ref{tab:SOA}) although 3FabRec was trained only with {\em 10\%} of the dataset. In addition, performance does not decrease much when going to lower values of 5\% and 1.5\% of training set size. Even when training with only {\em 10 images or 1 image}, our approach is able to deliver reasonably robust results (see Fig.\ref{fig:teaser} for landmark reconstruction results from training with 10 images). 

\vspace{-0.3cm}\paragraph{AFLW} For this dataset (Table \ref{tab:fewshotAFLW}), our approach already starts to come ahead at 20\% of training set size with little degradation down to 1\%. Again, even with only a few images 3FabRec can make landmark predictions. 

\vspace{-0.3cm}\paragraph{WFLW} For this more challenging dataset (Table \ref{tab:fewshotWFLW}), our approach easily outperforms the StyleAlign \cite{Honari2018} method as soon as less than 10\% is used for training while being able to maintain landmark prediction capabilities down to only 10 images in the training set.

\begin{table*}[ht]
	\footnotesize
	\begin{center}
		\begin{tabular}{p{1.47cm}| 
				p{0.31cm}p{0.31cm}| p{0.31cm}p{0.31cm}| p{0.31cm}p{0.31cm} | p{0.31cm}p{0.31cm}| p{0.31cm}p{0.31cm}| p{0.31cm}p{0.31cm}| p{0.31cm}p{0.31cm} | p{0.31cm}p{0.31cm}}
			\toprule
			\multicolumn{15}{c}{\bfseries  AFLW dataset}  \\
			\toprule
			\bfseries Method     &  \multicolumn{14}{c}{\bfseries  Training set size}  \\
			&    \multicolumn{2}{c}{100\%} &   \multicolumn{2}{c}{20\%}   & \multicolumn{2}{c}{10\%}    &   \multicolumn{2}{c}{5\%}    & \multicolumn{2}{c}{1\%} &   \multicolumn{2}{c}{50 (0.0025\%)} &  \multicolumn{2}{c}{10 (0.0005\%)} & \multicolumn{2}{c}{1 ($<$0.0001\%)} \\
			\toprule
			RCN$^+$ \cite{Honari2018} 	&\bf 1.61& - 	& - 	& - 	& - 	& - 	& 2.17 & - 	    &  2.88  & - 	& - 	& - 	& -  	& - 	& -    & -     \\  
			TS$^3$ \cite{Dong2019}    	&\bf - 	& - 	&   1.99 	& 1.86 	& 2.14 	& 1.94 	& 2.19 & 2.03 	&   -    &  -   & -     & - 	& - 	& -  	& - 	& -\\    
			\midrule
			3FabRec         			&   1.87 & \bf 1.59 &  \bf 1.96 & \bf 1.74 & \bf 2.03 & \bf 1.74 & \bf 2.13 & \bf 1.86 & \bf 2.38 & \bf 2.03 & \bf 2.74 & \bf 2.23  & \bf 3.05 & \bf 2.56 & \bf 4.93 & \bf 4.04\\  
			\midrule
			\bottomrule
		\end{tabular}	
	\end{center}
	\vspace{-0.5cm}
	\caption{\small NME (\%) with reduced training sets for AFLW. The first column in each cell is the full testset, the second is the frontal testset, "-" denotes values not reported.}	
	\label{tab:fewshotAFLW}
	\vspace{-0.0cm}
\end{table*}

\begin{table}
	\footnotesize
	\begin{center}
		\begin{tabular}{l| c | c | c | c | c |c | c }
			
			\toprule
			\multicolumn{7}{c}{\bfseries  WFLW dataset}  \\
			\toprule
			\bfseries Method     &  \multicolumn{7}{c}{\bfseries  Training set size}  \\
			&   100\%     &  20\%  &   10\%   &  5\% & 50 & 10  & 1  \\
			\toprule
			SA \cite{Honari2018} &   \bf 4.39  &  \bf 6.00  &   7.20   &  -    & -   &  -             \\
			\midrule
			3FabRec             &    5.62           &   6.51     & \bf 6.73  &  \bf 7.68   &  \bf 8.39    & \bf  9.66   & \bf 15.79     \\
			\midrule
			\bottomrule
		\end{tabular}	
	\end{center}
	\vspace{-0.5cm}
	\caption{\small NME (\%) with reduced training sets for WFLW.}	
	\label{tab:fewshotWFLW}
	\vspace{-0.0cm}
\end{table}

\subsection{Ablation studies}\label{sec:ablation}

\subsubsection{Effects of ITLs}
\begin{figure}
	\begin{center}
		\includegraphics[width=\columnwidth]{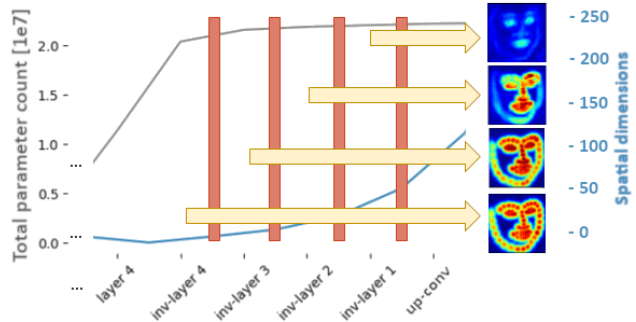}
	\end{center}
	\vspace{-0.5cm}
	\caption{\small Layer-analysis of 3FabRec. Gray curve: cumulative number of network parameters; blue curve: spatial dimension of each layer. The four red blocks indicate the ITL layers, with arrows showing how well the landmark heatmap can be predicted when starting from that layer.}
	\label{fig:ablation}
	\vspace{-0.2cm}
\end{figure}

\vspace{-0.1cm}
In order to see where information about landmarks is learned in the interleaved transfer layers, Figure \ref{fig:ablation} shows the reconstruction of the landmark heatmap when using all four layers versus decreasing subsets of the upper layers. As can be seen, the highest layer has only very localized information (mostly centered on eyes and mouth), whereas the lower layers are able to add information about the outlines - especially below layer 2. 

Localization accuracy is reported on the 300-W dataset (NME of 51 inner landmarks and outlines, as well as FR)  in Table \ref{tab:ablationITL}. As can be expected from the visualization, performance is bad for the upper layers only, but quickly recovers (especially when including the outlines) below layer 2. The reason for this is that the upper layers of the generator will mostly contain localized, de-correlated information at the pixel level, whereas the lower layers are closer to the more global and contextual information necessary to cover highly variable outlines (cf. blue curve in Figure \ref{fig:ablation}, note that all ITLs have 3$\times$3 convolutions). As the gray curve in Figure \ref{fig:ablation} and Table \ref{tab:ablationITL} show as well, the ITLs can achieve this with only very few additional parameters.

\vspace{-0.2cm}\subsubsection{Effects of finetuning}  
Table \ref{tab:ablationFT} reports the effects of running the model with and without finetuning on the full testsets of the three evaluated datasets. The additional retraining of the autoencoder allows for better reconstruction of the faces and results in benefits of 10.9\% on average (8.9\% for 300-W, 15.2\% for AFLW, and 8.5\% for WFLW, respectively).

\begin{table}
	\footnotesize
	\begin{center}
		\begin{tabular}{l | p{1cm} | p{1.2cm} |p{1cm} | p{1cm} }
			
			\toprule
			&   \multicolumn{4}{c}{\bfseries  Trained ITLs} \\
			& \bf 1+2+3+4  & \bf  2+3+4 &\bf  3+4  &  \bf 4 \\
			\toprule
			Input size              & 256x8x8 & 128x16x16 & 64x32x32 & 64x64x64 \\
			Trainable params  & 881k & 291k & 143k & 106k   \\
			300-W NME $\neg O$   &    3.54 & 3.63 & 5.34 & 16.34         \\
			300-W NME $O$    &       6.58 & 7.32  & 18.17 & 40.24       \\
			300-W FR@0.1    &  1.45      & 2.03  & 22.93 &  91.44            \\
			\midrule
			\bottomrule
		\end{tabular}	
	\end{center}
	\vspace{-0.5cm}
	\caption{\small Parameters and training results for ITLs ($\neg O$=without outlines, $O$=outlines only)}	
	\label{tab:ablationITL}
	\vspace{-0.1cm}
\end{table}

\begin{table}
	\footnotesize
	\begin{center}
		\begin{tabular}{l | c | c|  c }
			&  \bf 300-W & \bf AFLW & \bf WFLW \\
			\toprule
			NME before FT & 4.16    & 2.12 & 6.11  \\
			NME after FT  &  3.82    & 1.84 & 5.62   \\
			\bottomrule
		\end{tabular}	
	\end{center}
	\vspace{-0.5cm}
	\caption{\small NME (\%) before and after finetuning on full testsets. }
	\label{tab:ablationFT}
	\vspace{-0.3cm}
\end{table}

\vspace{-0.0cm}\subsection{Runtime performance}
Since inference complexity is equivalent to two forward-passes through a ResNet-18, our method is able to run at frame rates of close to 300fps on a TitanX GPU - an order of magnitude faster than state-of-the-art approaches with similar, high accuracy (LAB \cite{Wu2018}=16fps, Wing \cite{Feng2017}=30fps, Deep Regression \cite{Lv2017}=83fps, Laplace \cite{Robinson2019}=20fps).

\section{Conclusion}\vspace{-0.0cm}
With 3FabRec, we have demonstrated that an unsupervised, generative training on large amounts of faces captures implicit information about face shape, making it possible to solve landmark localization with only a minimal amount of supervised follow-up training. This paradigm makes our approach inherently more robust against overfitting to specific training datasets as well as against human annotation variability \cite{Dong2018}. The critical ingredients of 3FabRec that enable this generalization are the use of an adversarial autoencoder that reconstructs high-quality faces from a low-dimensional latent space, together with low-overhead, interleaved transfer layers added to the generator stage that transfer face reconstruction to landmark heatmap reconstruction. 

Results show that the autoencoder is easily able to generalize from its unlabeled training set to data from unseen datasets. This offers generalization for training from only a few percent of the training set and still produces reliable results from only a few annotated images - far below anything reported so far in the literature.
At the same time, since inference amounts to only two forward passes through a ResNet18, our method achieves much higher runtime performance than other highly accurate methods.


\paragraph{Acknowledgements}\vspace{-0.0cm}
This work was supported by Institute of Information \& Communications Technology Planning \& Evaluation (IITP) grant funded by the Korean government (MSIT) (No. 2019-0-00079, Department of Artificial Intelligence, Korea University)

\clearpage
{\small
    \bibliographystyle{ieee_fullname}
    \bibliography{cvpr2020}
}

\clearpage
\appendix

\begin{figure*}
	\begin{center}
         \includegraphics[width=1.0\linewidth]{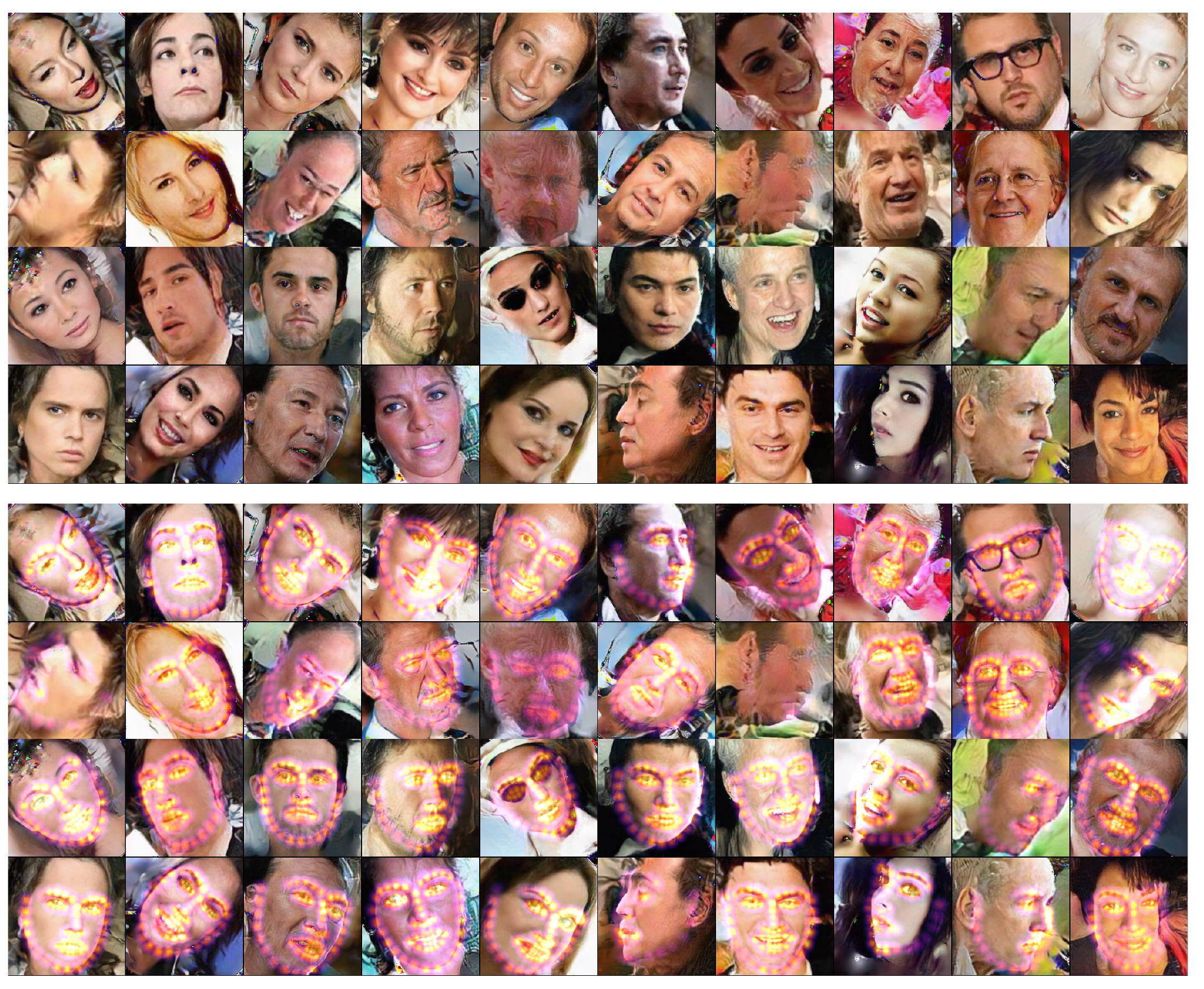}
	\end{center}
    \vspace{-0.4cm}
    \captionof{figure}{\small Randomly-generated faces from the 3FabRec framework (top four rows) together with their predicted landmark confidence heatmaps (bottom four rows).}
    \label{fig:random_faces}
\end{figure*}

\begin{table*}
	\small
	\begin{center}
		\begin{tabular}{cp{3.4cm}|p{0.4cm}p{0.4cm}p{0.5cm}|c|ccc|cc}
			\toprule
			& \bf Model& $\mathcal{L}_{rec}$ & $\mathcal{L}_{adv}$ & $\mathcal{L}_{cs}$ & FT & \multicolumn{2}{c}{\bf Global Reconstr.} & \bf Local Reconstr. & \bf NME & \bf FR@0.1 \\
			&  &                     &                     &                    &    &   RMSE & SSIM                                 &  Patch SSIM     &   \%  & \% (\#) \\
			\toprule
			(R)    & ResNet-18 &  &  &  & \checkmark  & -  & -  &  - & 5.64 & 4.64 (32) \\
			\midrule
			(HG) & Heatmap HG &  &  &  & \checkmark & - & - & - & 5.48 & 4.21 (29) \\
			\midrule
			\midrule
			(A)    & Adv. Autoencoder & \checkmark &  &  &  & \bf 12.61 & \bf 0.68 & \bf 0.64 & 5.67 & 4.94 (34) \\
			(A-FT) & Adv. Autoencoder (FT) & \checkmark &  &  & \checkmark & 25.03 & 0.57 & 0.55 & 4.92 & 2.47 (17) \\
			\midrule
			(B)    & AE + GAN & \checkmark & \checkmark &  &                 & 15.10 & 0.60 & 0.58 & 5.30 & 3.77 (26) \\
			(B-FT) & AE + GAN (FT) & \checkmark & \checkmark &  & \checkmark & 27.48 & 0.49 & 0.50 & 4.71 & 2.03 (14)  \\
			\midrule
			\midrule
			(C)    & AE + GAN + Struct. & \checkmark & \checkmark & \checkmark &                & 15.91 & 0.62 & 0.64 & 4.92 & 2.61 (18) \\
			(C-FT) & AE + GAN + Struct. (FT)& \checkmark & \checkmark & \checkmark & \checkmark & 27.65 & 0.50 & 0.53 & \bf 4.41 & \bf 1.45 (10) \\
			\bottomrule
		\end{tabular}	
	\end{center}
	\vspace{-0.4cm}
	\caption{Results of autoencoder ablation study. Rows (R) and (HG) are benchmark results from fully supervised methods with a comparable ResNet-18 architecture. Rows (A), (B), (C) show the effects of adding loss terms on both global and local reconstruction errors as well as on landmark localization accuracy and failure rate. Rows (A-FT), (B-FT), (C-FT) report results on post-finetuning the autoencoder on the 300-W dataset. NME = Normalized mean error, FR@0.1 = failure rate at 10\% NME. All results reported for the full testset of 300-W.}	
	\label{tab:ae_ablation}
\end{table*} 

\section{Random faces}\label{random} \vspace{-0.2cm}
Figure \ref{fig:random_faces} shows generated faces from a random sampling of the latent space (top four rows) together with the predicted landmark heatmaps (bottom four rows) using the final architecture from the main paper (trained on VGGFace2 and AffectNet with 256x256px). We note that the faces have high visual quality as well as large variability in facial appearance (pose, expression, hair style, accessories).

\section{Ablation studies}\label{ablation}

A critical part of our framework is the first step in which an adversarial autoencoder is trained in an unsupervised fashion on a large dataset of faces, which yields a low-dimensional embedding vector $z$ that encapsulates the face representation. 

\subsection{Autoencoder losses}\label{losses}

The adversarial autoencoder is trained through four loss functions balancing faithful image reconstruction with the generalizability and smoothness of the embedding space needed for the generation of novel faces. A reconstruction loss $\mathcal{L}_{rec}$ penalizes reconstruction errors through a pixel-based $L1$ error. An encoding feature loss $\mathcal{L}_{enc}$ \cite{goodfellow2014generative} ensures the creation of a smooth and continuous latent space. An adversarial feature loss  $\mathcal{L}_{adv}$ pushes the encoder $E$ and generator $G$ to produce reconstructions with high fidelity since training of generative model using only image reconstruction losses typically leads to blurred images. As the predicted landmark locations in our method follow directly from the locations of reconstructed facial elements, our main priority in training the autoencoder lies in the accurate reconstruction of such features, reconstruction accuracy is further enhanced by introducing a structural image loss  $\mathcal{L}_{cs}$. 

\begin{figure*}
	\begin{center}
		\includegraphics[width=1.00\textwidth]{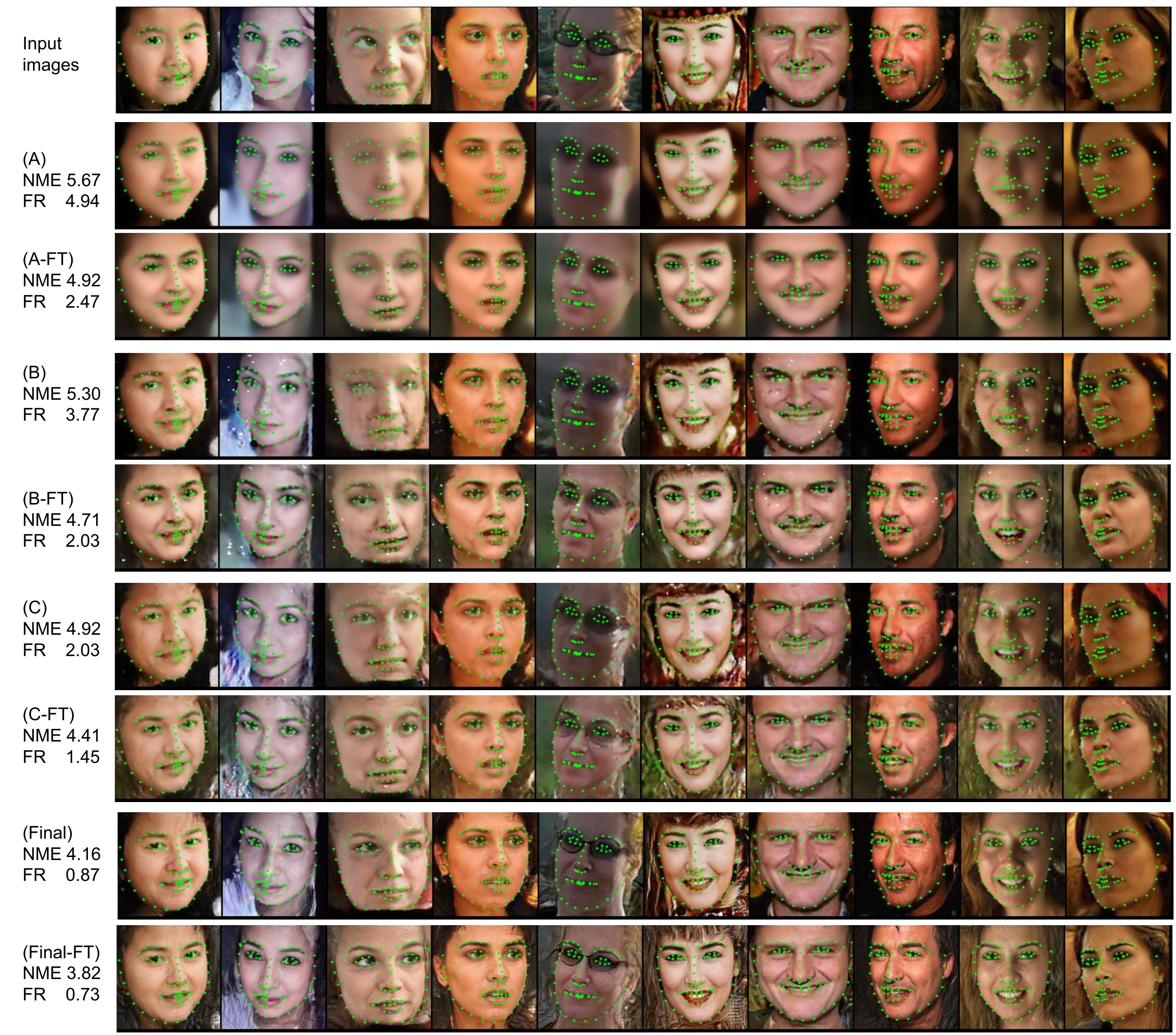}
	\end{center}
	\caption{Example reconstructions corresponding to Tab. \ref{tab:ae_ablation}. (A)-(C) are trained for 30 epoches on $128\times128$ images. 'Final' denotes the fully trained model $256\times256$ that was used for the experiments in Sec. 4 of the paper.}
	\label{fig:ae_ablation}
\end{figure*}

Here, we present results of the framework ablating different loss terms (except for the encoding feature loss $\mathcal{L}_{enc}$) during the training of the autoencoder to study their impact on landmark localization accuracy (see Table \ref{tab:ae_ablation}) using the 300-W dataset. In addition, we report the effects of the optional finetuning step on accuracy, in which the autoencoder is further tuned on the 300-W training dataset. All setups were trained on 128x128px images at a half of the resolution of the setup reported in the paper (see also Figure \ref{fig:ae_ablation}).

As benchmarks, the first two rows of Table \ref{tab:ae_ablation} also list a standard ResNet-18 predictor of landmark locations (trained on 300-W) as well as a standard heatmap-based system (trained on 300-W). Both approaches offer roughly the same kind of performance on this dataset with a slight advantage for heatmap-based prediction. 

If we only add the autoencoder (using  $\mathcal{L}_{rec}$,  $\mathcal{L}_{enc}$) to our ResNet-architecture, then performance is comparable to that of the standard, non-bottlenecked ResNet-18 architecture, which shows that the 99 dimensions seem to be sufficient to capture the landmark "knowledge" - it is important to note, however, that this landmark knowledge was obtained from {\em unsupervised} training. Further (supervised) finetuning of the autoencoder on 300-W provides another, significant boost that goes beyond the performance of both supervised benchmark systems. Hence, the finetuning step on the dataset is able to sharpen the implicit landmark representation obtained during the unsupervised step. 

Forcing the autoencoder to generate believable images by adding the adversarial loss (using  $\mathcal{L}_{rec}$,  $\mathcal{L}_{enc}$, $\mathcal{L}_{adv}$) provides a further 7\% improvement in NME for standard and finetuned training. Finally, the addition of the structural loss that further enhances small details in the reconstructed faces (using  $\mathcal{L}_{rec}$,  $\mathcal{L}_{enc}$, $\mathcal{L}_{adv}$, $\mathcal{L}_{cs}$)  yields another $\approx$7\% improvement. Overall, these results clearly show that losses that tune the face representation to be able to generate more detailed faces will also improve the landmark localization accuracy.

We note that the columns reporting "global" reconstruction errors (as RMSE or SSIM comparisons between the original and reconstructed images, respectively) and "local" reconstruction error (as SSIM errors evaluated for patches centered on the landmark locations of the original and reconstructed images) yield already good quality for the most "simple" loss setup. For this it is best to look at Figure \ref{fig:ae_ablation}, which shows how the different losses affect the visual quality of the reconstruction. When looking at rows (A), (B), (C), faces gain an increasing amount of high-frequency detail. When adding the GAN loss, these high-frequency details will not aid the reconstruction error at first as the details are "hallucinated" globally all over the face - these details, however, seem to be able to aid the landmark layers in providing a better mapping onto heatmaps and therefore landmark locations. The addition of the SSIM loss does improve the reconstruction error again as the loss forces the high-frequency details to better match with the trained source face images - again, the added details in this case will help landmark localization.

The effect of finetuning on face appearance is interesting to observe as the faces gain immediate detail for all loss setups, yet their overall reconstruction is sometimes more "different" to the source face compared to the non-finetuned version. This is because finetuning unfreezes the weights of the encoder but will train to predict the landmark locations more reliably - hence, the reconstructed faces will favor clear landmark localizability (through well-defined facial feature locations) at the expense of more faithful face reconstruction. Overall, the effect is therefore an increase of the reconstruction error. 

As a final note, we observe that training the autoencoder setup on 256x256px provides another jump in performance as the system will learn to reconstruct facial details at an even higher fidelity (see final two rows in Figure \ref{fig:ae_ablation}).

\subsection{Encoding length}\label{enclength}
\begin{table}
	\small
	\begin{center}
		\begin{tabular}{c|cc}
			\toprule
			\bf \# Dims& \bf NME & \bf FR@0.1 \\
			& \% &  \% (\#) \\
			
			\toprule
			50$^\dag$ & 4.59 & \bf 1.02 (07) \\
			99 & \bf 4.41 & 1.45 (10)  \\
			\bottomrule
		\end{tabular}	
	\end{center}
	\vspace{-0.4cm}
	\caption{Number of dimension of embedded feature vectors. $^\dag$ Landmark training was instable and required multiple restarts and a reduction of the learning rate.  }	
	\label{tab:enclength}
\end{table}

The latent vector $z$ reported in the main paper has a dimensionality of $d=99$ which is comparable to other GAN-frameworks \cite{schroff2015facenet,radford2015unsupervised}. 

In Table \ref{tab:enclength}, we report the effect of halving this dimensionality to $d=50$ on landmark localization accuracy. Although yielding a slightly higher NME, the reduced autoencoder obtains a slightly lower FR, which overall means that both embedding dimensionalities result in similar performance levels. An issue with the reduced dimensionality embedding, however, was that the subsequent landmark training was notably less robust, requiring a much more conservative learning rate.

Hence, for the task of landmark localization, the current framework may work with a lower-dimensional embedding space, however, it seems that pulling the implicit information out of the reduced dimensions is a harder task than for a richer embedding.

Further experiments are needed to investigate the effects of increasing the dimensionality as well as providing further constraints on the embedding vector $z$ during the unsupervised training.

\begin{table*}
	\begin{center}
		\begin{tabular}{lclcc|rr|rr}
			\toprule
			& & \multicolumn{3}{c}{\bfseries Unlabeled training data}  &  \multicolumn{4}{c}{\bfseries Labeled training data} \\
			\toprule
			\bf Model & \bf Num. & \bf Pre-train & \bf Num. of & \bf External  & \multicolumn{2}{c}{\bfseries 100\% (3,189)} & \multicolumn{2}{c}{\bfseries 1.5\% (50)} \\
			& \bf param. & \bf dataset(s) & \bf images & \bf images & NME & FR@0.1 &  NME &  FR@0.1 \\
			\toprule
			ResNet-18 & 11M & None & 0 & no & 5.64 & 4.64 (32) & 8.70 & 22.21 (153)  \\
			Heatmap HG & 22M & None & 0 & no & 5.48 & 4.21 (29) & 10.13 & 39.33 (271)  \\
			\midrule
			C-FT & 23M & 300-W & 3,189 & no & 5.40 & 4.79 (33) & 7.95& 15.82 (109)\\
			\midrule
			C-FT & 23M & VGG + AN & 100k & yes & 4.73 & 1.74 (12) & 6.34 & 9.29 (064) \\
			C-FT & 23M & VGG + AN & 2.1M & yes & \bf 4.41 & \bf 1.45 (10) & \bf 5.71 & \bf 4.35 (030) \\
			\bottomrule
		\end{tabular}	
	\end{center}
	\vspace{-0.4cm}
	\caption{Effect of unsupervised pre-training when trained with full and reduced labeled training data on 300-W.}	
	\label{tab:training}
\end{table*}

\subsection{Unsupervised training and few-shot learning}\label{training}

We next take a look at the effects of the unsupervised training step as well as the amount of supervised post-training on 300-W. Table \ref{tab:training} shows again the ResNet-18 and heatmap hourglass baselines and then three different training setups for our full, finetuned system at 128x128px image size. 

The first two rows report results of the full architecture without any unsupervised pre-training and hence without any implicit face knowledge. 
The next rows show results for the full architecture with different amounts of pre-training. Pre-training on the 300-W training dataset results in equal or slightly better performance compared to the baseline architectures showing that the system is able to pick up implicit knowledge already from only 3,200 images. Pre-training on 100,000 images provides a significant, further jump as does pre-training on the full 2,1M image dataset.

Importantly, the error increase in the presence of limited training data (columns labeled 1.5\% in Table \ref{tab:training}) with just 50 images showcase the power of the pre-trained representation: whereas ResNet-18 increases around 54\% in NMW from 100\% to 1.5\% training set size, our pre-trained architectures only reduce 47\%, 34\%, and 29\% respectively owing to the more robust generalization from the latent representation.

\section{Few-shot learning on different datasets}\label{fewshot}

\begin{figure*}
	\begin{center}
		\includegraphics[width=1.00\textwidth]{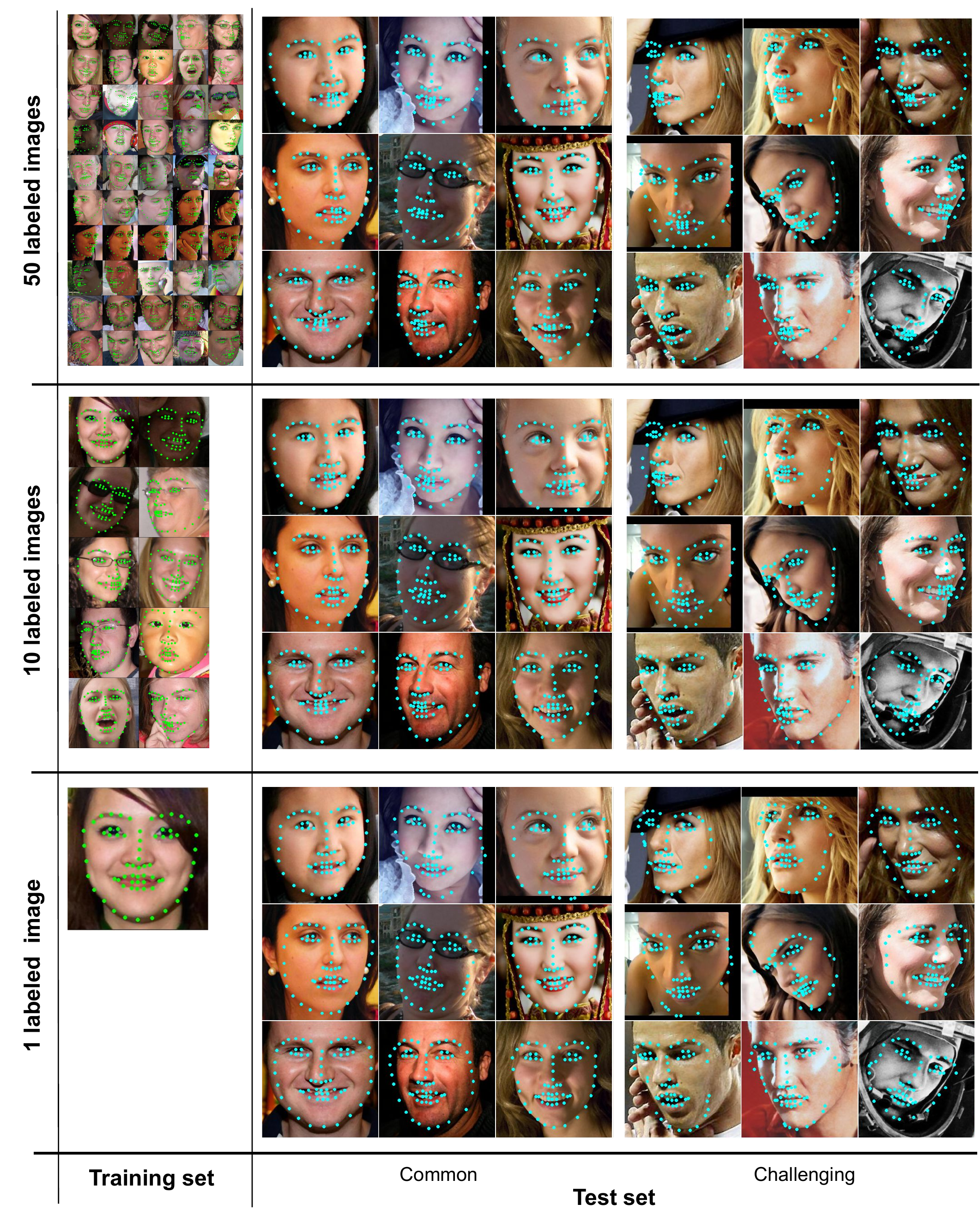}
	\end{center}
	\caption{Few-shot learning on 300-W}
	\label{fig:fs_300w}
\end{figure*}

\begin{figure*}
	\begin{center}
		\includegraphics[width=1.00\textwidth]{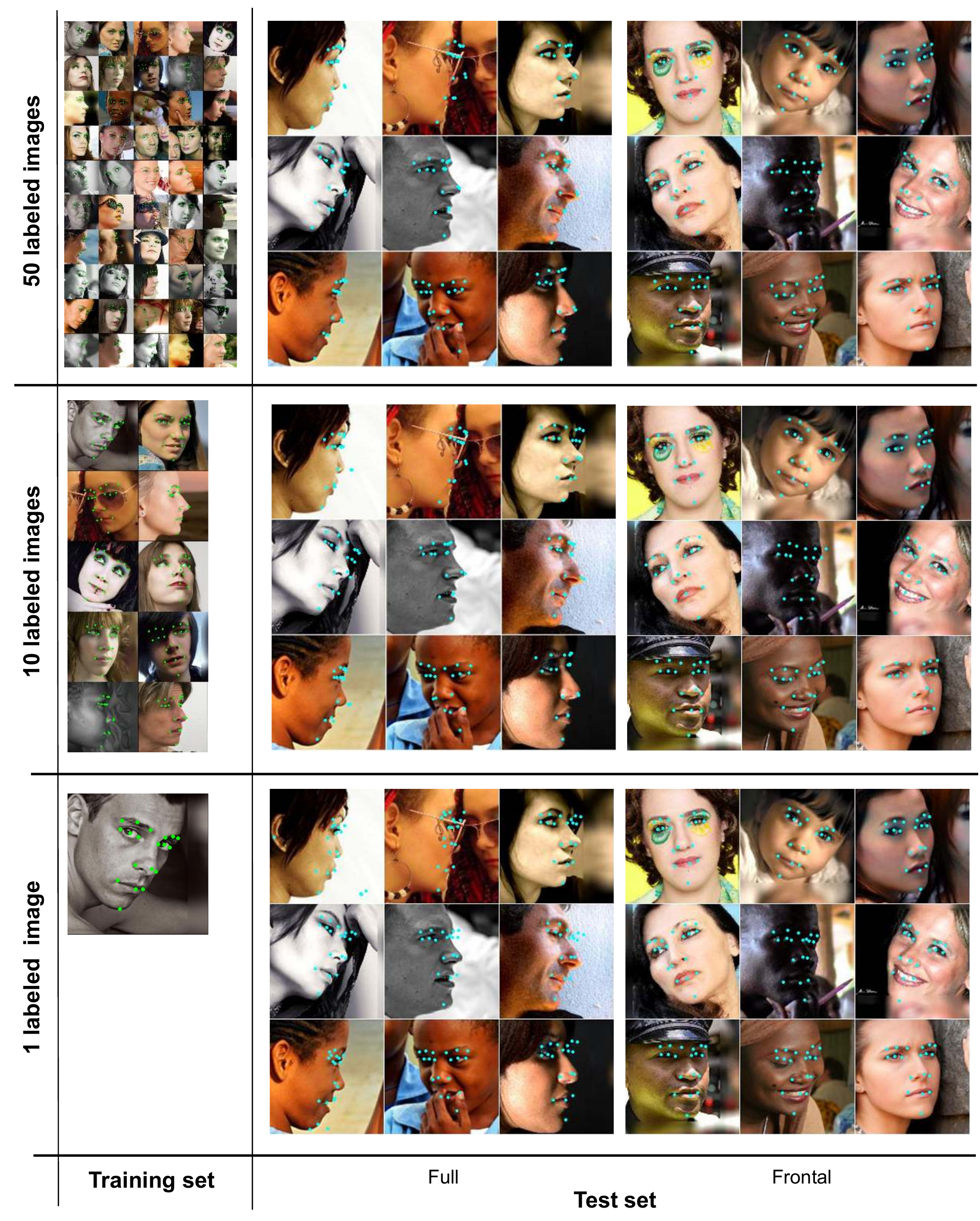}
	\end{center}
	\caption{Few-shot learning on AFLW}
	\label{fig:fs_aflw}
\end{figure*}

\begin{figure*}
	\begin{center}
		\includegraphics[width=1.00\textwidth]{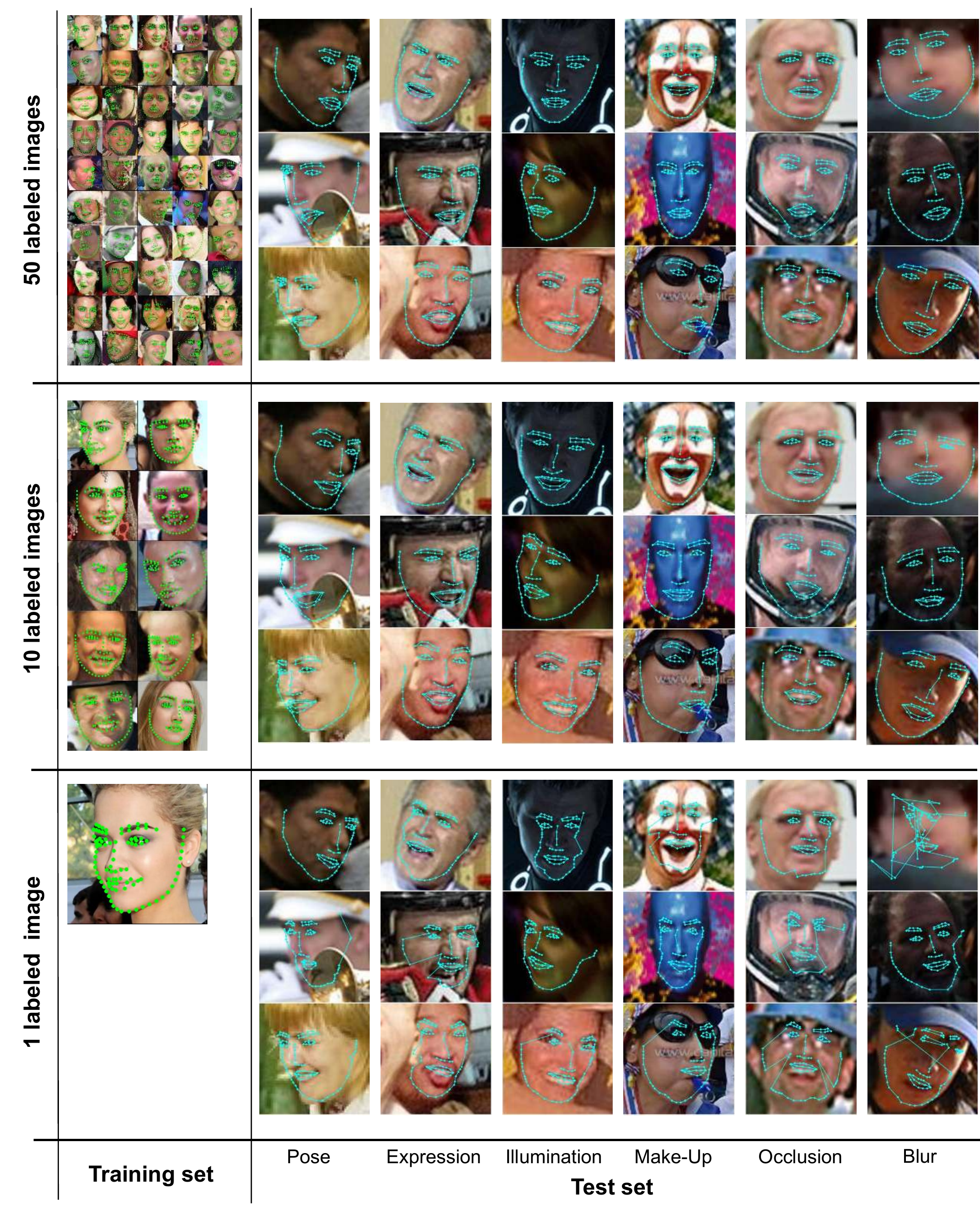}
	\end{center}
	\caption{Few-shot learning on WFLW}
	\label{fig:fs_wflw}
\end{figure*}

Figures \ref{fig:fs_300w},\ref{fig:fs_aflw},\ref{fig:fs_wflw} show results for few-shot learning on the three different datasets (300-W, AFLW, WFLW) reported in the main paper. The first column has the {\em entire} training set (50, 10, or 1 labeled image(s)), and the second column shows predicted landmarks on nine or three images from the different testsets contained in the datasets. In all figures, training with even just one image produces reasonable localization results and a clear improvement in prediction accuracy can be traced as a few more images are added.

In Figure \ref{fig:fs_wflw}, the failure cases are most visible (see, for example, the top results for training with one image on the Blur testset). It should be noted that this is by far the most challenging dataset as it contains variability in face appearance (due to illumination, occlusion, and make-up) that is not fully present in the unsupervised datasets we used (cf. the randomly-generated faces in Figure \ref{fig:random_faces}). As a few more labeled faces are added, however, performance begins to quickly improve even in the presence of such severe changes.

\section{Additional result visualizations}\label{added}
\begin{figure*}
	\begin{center}
		\includegraphics[width=0.94\textwidth]{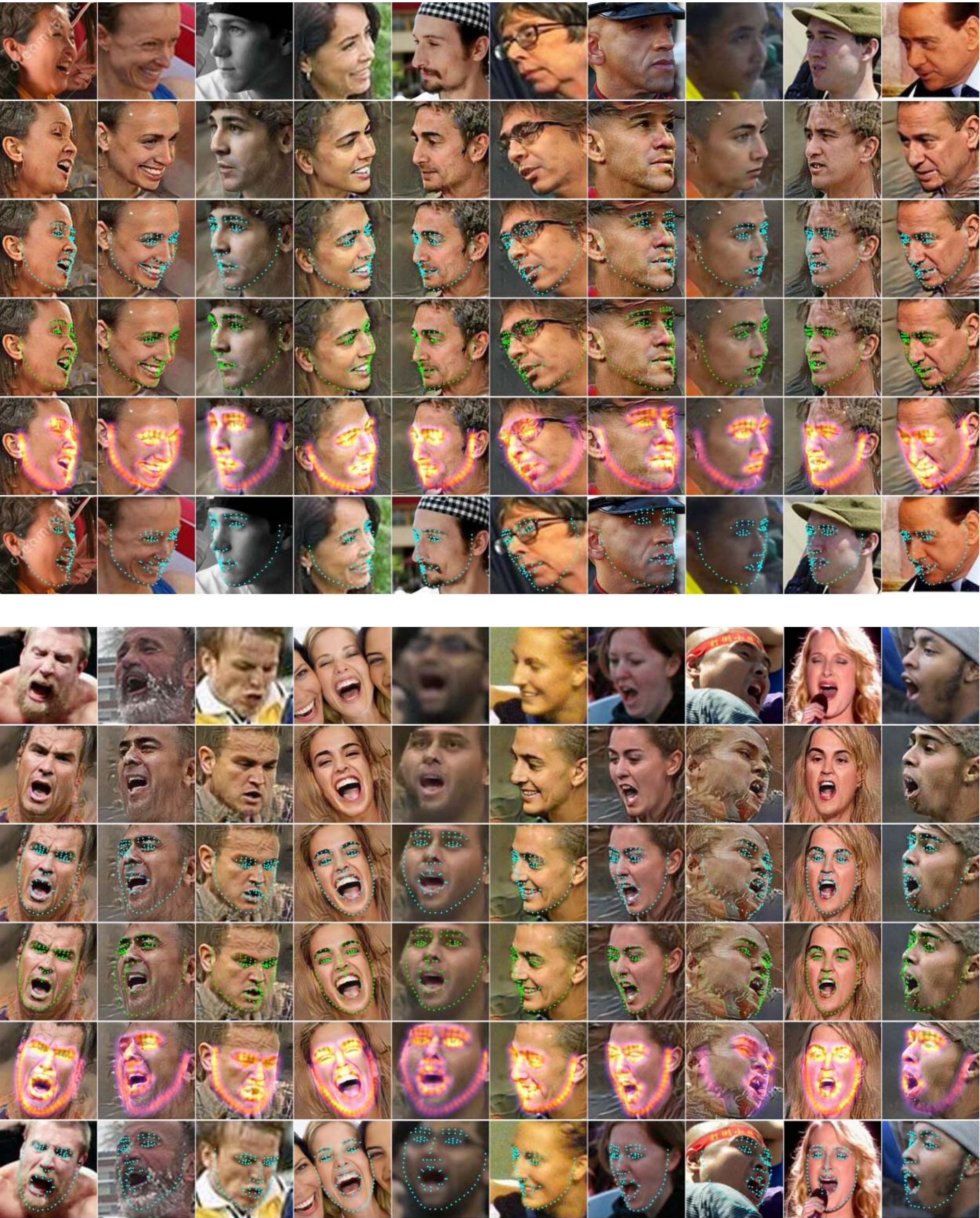}
	\end{center}
	\caption{3FabRec results on WFLW Pose and Expression: Two blocks of rows show (1) original, (2) reconstruction, (3) reconstruction with predicted landmarks,  (4) reconstruction with ground-truth landmarks, (5) reconstruction with predicted landmark heatmaps, and (6) original with predicted landmarks, respectively.}
	\label{fig:sample_300w}
\end{figure*}

\begin{figure*}
	\begin{center}
		\includegraphics[width=0.94\textwidth]{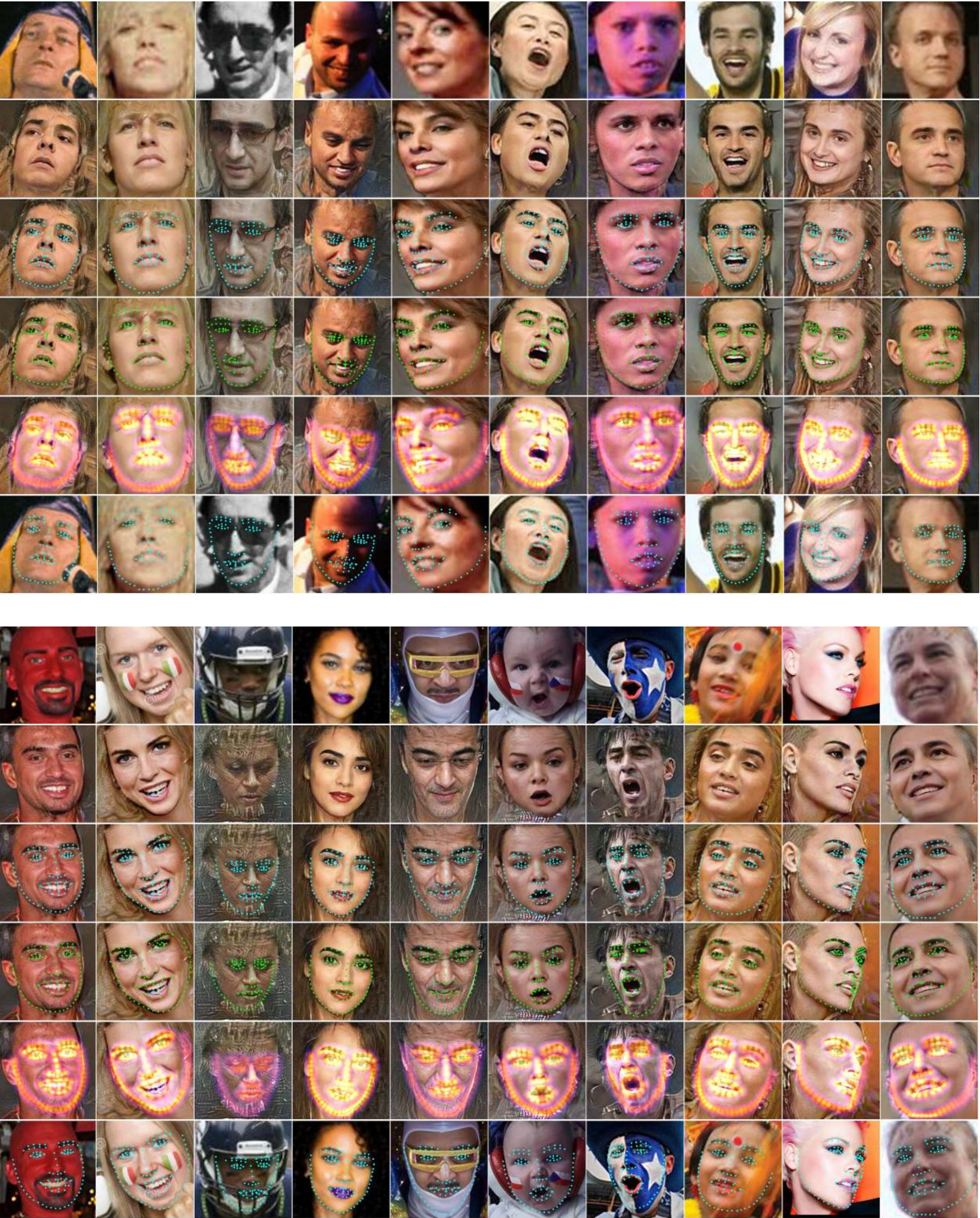}
	\end{center}
	\caption{3FabRec results on WFLW Illumination and Make-Up: Two blocks of rows  show (1) original, (2) reconstruction, (3) reconstruction with predicted landmarks,  (4) reconstruction with ground-truth landmarks, (5) reconstruction with predicted landmark heatmaps, and (6) original with predicted landmarks, respectively.}
	\label{fig:sample_aflw}
\end{figure*}

\begin{figure*}
	\begin{center}
		\includegraphics[width=0.94\textwidth]{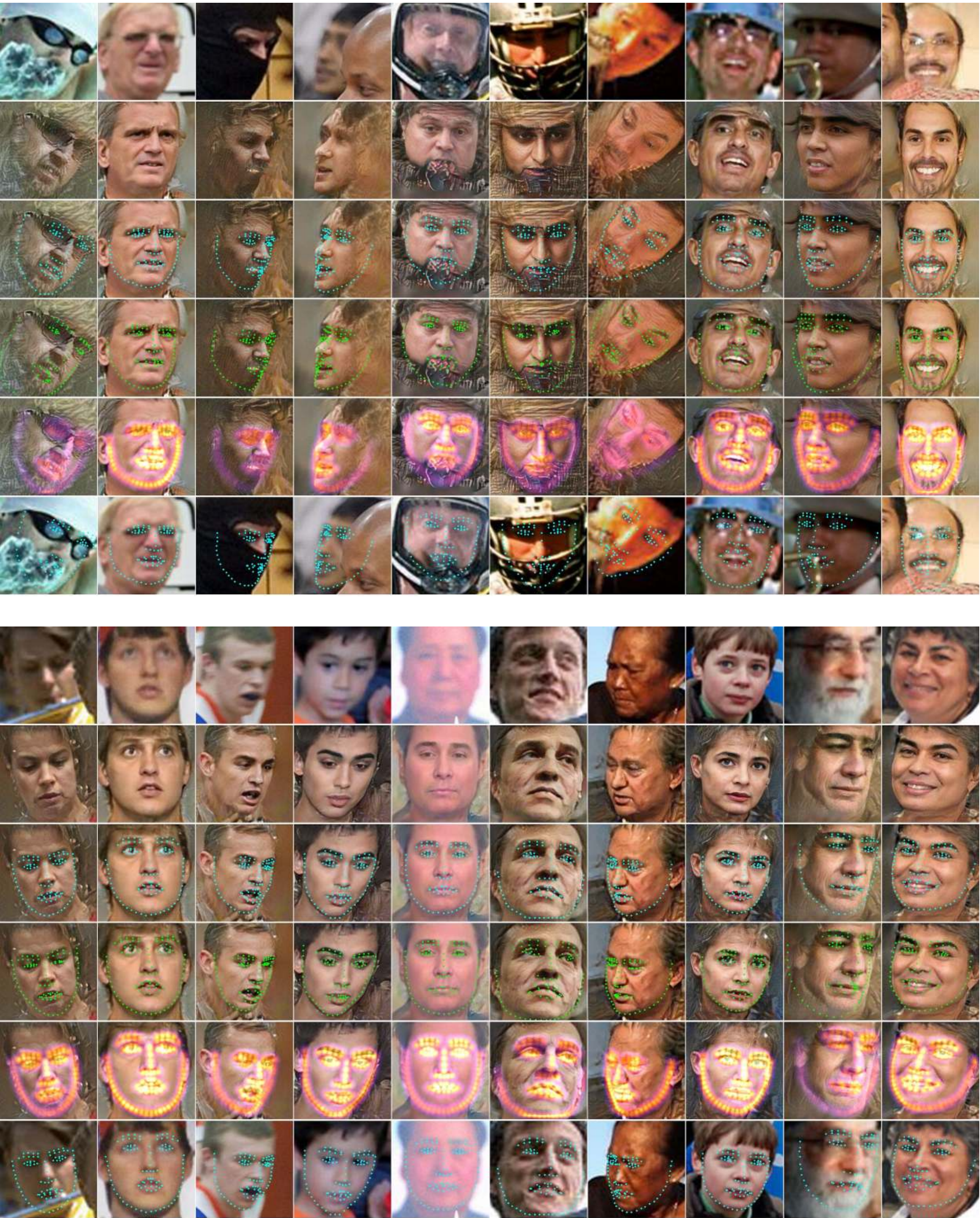}
	\end{center}
	\caption{3FabRec results on WFLW Occlusion and Blur: Two blocks of rows  show (1) original, (2) reconstruction, (3) reconstruction with predicted landmarks,  (4) reconstruction with ground-truth landmarks, (5) reconstruction with predicted landmark heatmaps, and (6) original with predicted landmarks, respectively.}
	\label{fig:sample_wflw}
\end{figure*}

Figures \ref{fig:sample_300w},\ref{fig:sample_aflw},\ref{fig:sample_wflw} show additional, non-curated visualizations of the full system on images from the six test subsets of WFLW.

\end{document}